\documentclass{article}
\PassOptionsToPackage{numbers, compress}{natbib}
\usepackage[final]{neurips_data_2023}
\usepackage[subfig,neurips]{definition}
\usepackage{arydshln}
\usepackage{booktabs}
\usepackage{float}
\usepackage[title]{appendix}
\usepackage{fontawesome5}
\usepackage{bbding}
\usepackage{pifont}

\newcommand{\myblue}[1]{\textcolor[RGB]{51,48,228}{\textbf{#1}}}
\newcommand{\mygreen}[1]{\textcolor[RGB]{54,174,124}{\textbf{#1}}}
\newcommand{\myorange}[1]{\textcolor[RGB]{248,111,3}{\textbf{#1}}}
\newcommand{\secref}[1]{Section~\ref{#1}}
\newcommand{\appref}[1]{Appendix~\ref{#1}}
\newcommand{\figref}[1]{Figure~\ref{#1}}
\newcommand{\tabref}[1]{Table~\ref{#1}}
\newcommand{\best}{\cellcolor[HTML]{B8E7E1}}
\newcommand{\second}{\cellcolor[HTML]{D4FAFC}}

\annotator{yanqiao}{red}

\title{GSLB: The Graph Structure Learning Benchmark}

\author{%
    \normalsize
    Zhixun Li\textsuperscript{\textmd{1}}\enspace
    Liang Wang\textsuperscript{\textmd{2,3}}\enspace
    Xin Sun\textsuperscript{\textmd{4}}\enspace
    Yifan Luo\textsuperscript{\textmd{5}}\enspace
    Yanqiao Zhu\textsuperscript{\textmd{6}}\enspace
    Dingshuo Chen\textsuperscript{\textmd{2,3}}\\
    \normalsize
    \textbf{Yingtao Luo}\textsuperscript{\textmd{7}}\enspace
    \textbf{Xiangxin Zhou}\textsuperscript{\textmd{2,3}}\enspace
    \textbf{Qiang Liu}\textsuperscript{\textmd{2,3}}\enspace
    \textbf{Shu Wu}\textsuperscript{\textmd{2,3}}\enspace
    \textbf{Liang Wang}\textsuperscript{\textmd{2,3,4}}\enspace
    \textbf{Jeffrey Xu Yu}\textsuperscript{\textmd{1}}\\
    \textsuperscript{\textmd{1}} Department of Systems Engineering and Engineering Management\\ The Chinese University of Hong Kong\\
    \textsuperscript{\textmd{2}}Center for Research on Intelligent Perception and Computing\\ State Key Laboratory of Multimodal Artificial Intelligence Systems\\ Institute of Automation, Chinese Academy of Sciences\\
    \textsuperscript{\textmd{3}} School of Artificial Intelligence, University of Chinese Academy of Sciences \\
    \textsuperscript{\textmd{4}}Department of Automation, University of Science and Technology of China\\
    \textsuperscript{\textmd{5}}School of Cyberspace Security, Beijing University of Posts and Telecommunications\\
    \textsuperscript{\textmd{6}}Department of Computer Science, University of California, Los Angeles\\
    \textsuperscript{\textmd{7}}Heinz College of Information Systems and Public Policy, Machine Learning Department \\
    School of Computer Science, Carnegie Mellon University\\
    \normalsize\rule{0pt}{1em}
    \faEnvelope[regular]{} Primary contact: \texttt{zxli@se.cuhk.edu.hk}\\
}

\begin{document}

\maketitle

\begin{abstract}
    Graph Structure Learning (GSL) has recently garnered considerable attention due to its ability to optimize both the parameters of Graph Neural Networks (GNNs) and the computation graph structure simultaneously. Despite the proliferation of GSL methods developed in recent years, there is no standard experimental setting or fair comparison for performance evaluation, which creates a great obstacle to understanding the progress in this field. To fill this gap, we systematically analyze the performance of GSL in different scenarios and develop a comprehensive \underline{G}raph \underline{S}tructure \underline{L}earning \underline{B}enchmark (GSLB) curated from 20 diverse graph datasets and 16 distinct GSL algorithms. Specifically, GSLB systematically investigates the characteristics of GSL in terms of three dimensions: \myblue{effectiveness}, \myorange{robustness}, and \mygreen{complexity}. We comprehensively evaluate state-of-the-art GSL algorithms in node- and graph-level tasks, and analyze their performance in robust learning and model complexity. Further, to facilitate reproducible research, we have developed an easy-to-use library for training, evaluating, and visualizing different GSL methods. Empirical results of our extensive experiments demonstrate the ability of GSL and reveal its potential benefits on various downstream tasks, offering insights and opportunities for future research. The code of GSLB is available at: \url{https://github.com/GSL-Benchmark/GSLB}.
\end{abstract}

\section{Introduction}

Graphs, structures made of vertices and edges, are ubiquitous in real-world applications. A wide variety of applications spanning social network\cite{zhang2022robust,fan2019graph}, molecular property prediction\cite{wieder2020compact,hao2020asgn}, fake news detection\cite{xu2022evidence,bian2020rumor}, and fraud detection\cite{li2022devil,liu2021pick} have found graphs instrumental in modeling complex systems. In recent years, Graph Neural Networks (GNNs) have attracted increasing attention due to their powerful ability to learn node or graph representations. However, most of the GNNs heavily rely on the assumption that the initial structure of the graph is trustworthy enough to serve as ground-truth for training. Due to uncertainty and complexity in data collection, graph structures are inevitably redundant, biased, noisy, incomplete, or the original graph structures are even unavailable, which will bring great challenges for the deployment of GNNs in real-world applications. 

To mitigate the aforementioned problems, Graph Structure Learning (GSL) \cite{chen2020iterative, zheng2020robust, luo2021learning, fatemi2021slaps, zhu2021survey, zhang2021mining} has become an important theme in graph learning.
GSL aims to make the computation structure of GNNs more suitable for downstream tasks and improve the quality of the learned representations. While it is widespread in different communities and the research enthusiasm for GSL is increasing, there is no standardized benchmark that could offer a fair and consistent comparison of different GSL algorithms. Moreover, due to the complexity and diversity of graph datasets, the experimental setups in existing work are not consistent, such as varying ratios of the training set and different train/validation/test splits. This poses a great obstacle to a holistic understanding of the current research status. Therefore, the development of a standardized and comprehensive benchmark for GSL is an urgent need within the community.

\begin{table}[]
\caption{An overview of GSLB. Both algorithms and datasets are divided into three categories: homogeneous node-level, heterogeneous node-level, and graph-level. The evaluation is divided into three dimensions: \myblue{effectiveness}, \myorange{robustness}, and \mygreen{complexity}.}
\begin{tabular}{ll}
\toprule
\multicolumn{2}{c}{\cellcolor[HTML]{C5DFF8}{\textbf{\emph{Algorithms}}}}                                                                             \\ \midrule[1pt]
Homogeneous GSL        & \begin{tabular}[c]{@{}l@{}} LDS\cite{franceschi2019learning}, GRCN\cite{yu2021graph}, ProGNN\cite{jin2020graph}, IDGL\cite{chen2020iterative}, CoGSL\cite{liu2022compact}, SUBLIME\cite{liu2022towards}, \\
GEN\cite{wang2021graph}, STABLE\cite{li2022reliable}, NodeFormer\cite{wu2022nodeformer}, SLAPS\cite{fatemi2021slaps},
GSR\cite{zhao2023self},
HES-GSL\cite{wu2023homophily}\end{tabular}  \\ \midrule
Heterogeneous GSL      & GTN\cite{yun2019graph}, HGSL\cite{zhao2021heterogeneous}                                                                         \\ \midrule
Graph-level GSL        & HGP-SL\cite{zhang2019hierarchical}, VIB-GSL\cite{sun2022graph}                                                                   \\ \midrule[1pt]
\multicolumn{2}{c}{\cellcolor[HTML]{C5DFF8}{\textbf{\emph{Datasets}}}}                                                                               \\ \midrule[1pt]
Homogeneous datasets   & \begin{tabular}[c]{@{}l@{}}Cora\cite{yang2016revisiting}, Citeseer\cite{yang2016revisiting}, Pubmed\cite{yang2016revisiting}, ogbn-arxiv\cite{hu2020open}, Polblogs, Cornell\cite{pei2020geom}, \\
Texas\cite{pei2020geom}, Wisconsin\cite{pei2020geom}, Actor\cite{tang2009social}\end{tabular}    \\ \midrule
Heterogeneous datasets & ACM\cite{yun2019graph}, DBLP\cite{yun2019graph}, Yelp\cite{lu2019relation}                                                                   \\ \midrule
Graph-level datasets   & \begin{tabular}[c]{@{}l@{}}IMDB-B\cite{cai2018simple}, IMDB-M\cite{cai2018simple}, COLLAB\cite{yanardag2015deep}, REDDIT-B\cite{yanardag2015deep}, MUTAG\cite{debnath1991structure}, \\ PROTEINS\cite{borgwardt2005protein}, Peptides-Func\cite{dwivedi2022long}, Peptides-Struct\cite{dwivedi2022long}\end{tabular} \\ 
\midrule[1pt]
\multicolumn{2}{c}{\cellcolor[HTML]{C5DFF8}{\textbf{\emph{Evaluations}}}}         \\ 
\midrule[1pt]
Effectiveness  & \begin{tabular}[c]{@{}l@{}}Homogeneous node classification (Topology Refinement/Topology Inference), \\
Heterogeneous node classification, Graph-level tasks\end{tabular} \\ \midrule
Robustness  & Supervision signal robustness, Structure robustness, Feature robustness \\ \midrule
Complexity  & Time complexity, Space complexity \\
\bottomrule
\end{tabular}
\label{tab:overview}
\end{table}

In this work, we propose \underline{G}raph \underline{S}tructure \underline{L}earning \underline{B}enchmark (GSLB), which serves as the first comprehensive benchmark for GSL. Our benchmark encompasses 16 state-of-the-art GSL algorithms and 20 diverse graph datasets covering homogeneous node-level, heterogeneous node-level, and graph-level tasks.
We systematically investigate the characteristics of GSL in terms of three dimensions: \myblue{effectiveness}, \myorange{robustness}, and \mygreen{complexity}.
Based on these three dimensions, we conduct an extensive comparative study of existing GSL algorithms in different scenarios.
For \myblue{effectiveness}, GSLB provides a fair and comprehensive comparison of existing algorithms on homogeneous node-level, heterogeneous node-level, and graph-level tasks, where we consider both homophilic and heterophilic graph datasets for homogeneous node-level tasks, and cover both Topology Refinement (TR, i.e., refining graphs from data with the original topology) and Topology Inference (TI, i.e., inferring graphs from data without initial topology) settings.
For \myorange{robustness}, GSLB evaluates GSL models under three types of noise: supervision signal noise, structure noise, and feature noise. We also compare GSL algorithms with the models specifically designed to improve these types of robustness.
For \mygreen{complexity}, GSLB conducts a detailed evaluation of representative GSL algorithms in terms of time complexity and space complexity.

Through extensive experiments, we observe that: (1) GSL generally brings performance improvement for node-level tasks, especially on heterophilic graphs; (2) on graph-level tasks, current GSL models bring limited improvement and their performance varies greatly across different datasets; (3) most GSL algorithms (especially unsupervised GSL algorithms) show impressive robustness; (4) GSL models require significant time and memory overhead, making them challenging to deploy on large-scale graphs.
In summary, we make the following three contributions:
\begin{itemize}
    \item We propose GSLB, the first comprehensive benchmark for graph structure learning. We integrate 16 state-of-the-art GSL algorithms and 20 diverse graph datasets covering homogeneous node-level, heterogeneous node-level, and graph-level tasks. An overview of our benchmark is shown in \tabref{tab:overview}.
    \item To explore the ability and limitations of GSL, we systematically evaluate existing algorithms from three dimensions: \myblue{effectiveness}, \myorange{robustness}, and \mygreen{complexity}. Based on the results, we reveal the potential benefits and drawbacks of GSL to assist future research efforts. 
    \item To facilitate future work and help researchers quickly use the latest models, we develop an easy-to-use open-source library. Besides, users can evaluate their own models or datasets with less effort. Our code is available at \url{https://github.com/GSL-Benchmark/GSLB}.
\end{itemize}

\section{Problem Definition}
\label{sec:definition}

In this section, we will briefly review the advances and basic concepts of GSL. Given an undirected graph $\mathcal{G}=(\mathbf{A},\mathbf{X})$, where $\mathbf{A}\in\mathbb{R}^{N\times N}$ is the adjacency matrix, $a_{uv}=1$ if edge $(u, v)$ exists and $a_{uv}=0$ otherwise, and $\mathbf{X}\in\mathbb{R}^{N\times F}$ is the node features matrix, $N$ is the number of nodes, $F$ is the dimension of node features. Given an optional graph $\mathcal{G}$, the goal of GSL is to jointly optimize computation graph $\mathcal{G}^\star=(\mathbf{A}^\star,\mathbf{X})$ and the parameters of graph encoder $\Theta_f$ to obtain high-quality node representations $\mathbf{Z}^\star\in\mathbb{R}^{N\times F'}$ for downstream tasks, where $\mathbf{A}^\star$ is the refined graph by graph learner.

\begin{figure}[t]
\centering
\includegraphics[width=\textwidth]{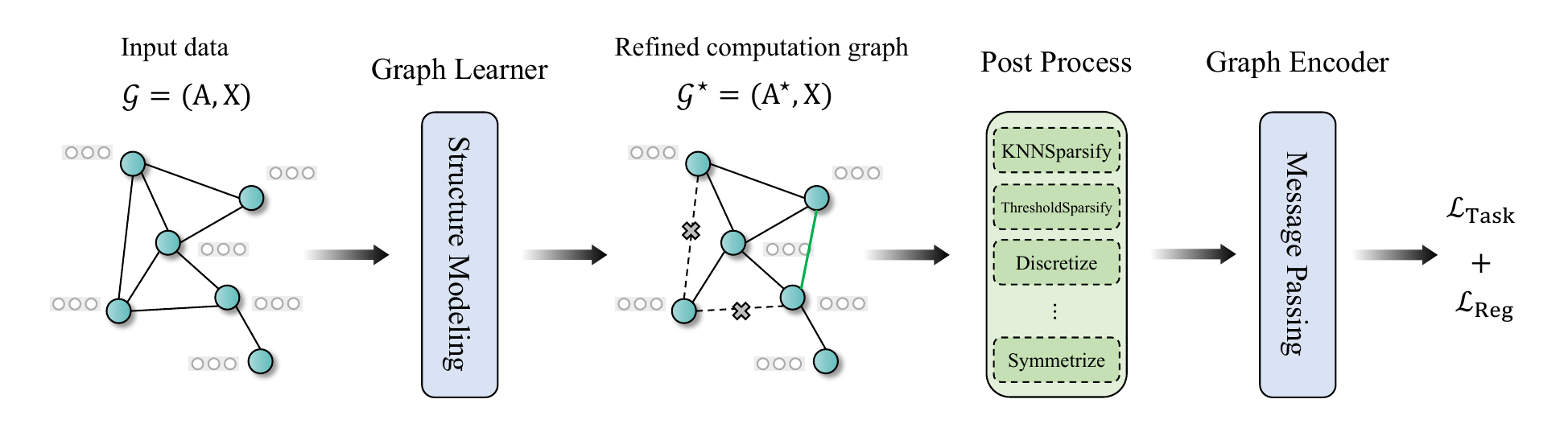}
\caption{A general framework of Graph Structure Learning (GSL). GSL methods start with input features and an optional initial graph structure. Its corresponding computation graph is refined/inferred through a structure learning module. With the learned computation graph, Graph Neural Networks (GNNs) are used to generate graph representations.}
\label{fig:framework}
\end{figure}

In general, the objective of GSL can be summarized as the following formula:
\begin{equation}
    \label{equ:objective}
    \mathcal{L}_{\text{GSL}}=\mathcal{L}_{\text{Task}}(\mathbf{Z}^\star,\mathbf{Y})+\lambda\mathcal{L}_{\text{Reg}}(\mathbf{A}^\star,\mathbf{Z}^\star,\mathcal{G})
\end{equation}
where the first term $\mathcal{L}_{\text{Task}}$ refers to a task-specific objective with respect to the learned representation $\mathbf{Z}^\star$ and ground-truth $\mathbf{Y}$, the second term $\mathcal{L}_{\text{Reg}}$ imposes constraints on the learned graph structure and representations, and $\lambda$ is a hyper-parameter that controls the trade-off between the two terms. The general framework of GSL is shown in \figref{fig:framework}.

\section{GSLB: Graph Structure Learning Benchmark}

In this section, we introduce the overview of \underline{G}raph \underline{S}tructure \underline{L}earning \underline{B}enchmark, with considerations of algorithms (\secref{sec:algorithms}), datasets (\secref{sec:datasets}) and evaluations (\secref{sec:evaluations}).

\subsection{Benchmark Algorithms}
\label{sec:algorithms}

\tabref{tab:overview} shows the overall 16 algorithms integrated in GSLB. They are divided into three categories: homogeneous GSL, heterogeneous GSL, and graph-level GSL. We briefly introduce each category in the following, and more details are provided in \appref{appen:alg}.

\textbf{Homogeneous GSL.} Most of the existing GSL algorithms are designed for homogeneous graphs. They assume there is only one type of nodes and edges in the graph. We select 7 TR-oriented algorithms including GRCN\cite{yu2021graph}, ProGNN\cite{jin2020graph}, IDGL\cite{chen2020iterative}, GEN\cite{wang2021graph}, CoGSL\cite{liu2022compact}, STABLE\cite{li2022reliable}, and GSR\cite{zhao2023self}. For TI-oriented algorithms, we select SUBLIME\cite{liu2022towards}, NodeFormer\cite{wu2022nodeformer}, SLAPS\cite{fatemi2021slaps}, and HES-GSL\cite{wu2023homophily}. It is worth noting that TR-oriented algorithms can only be applied if the original graph structure is available, but we can construct a preliminary graph based on node features (e.g., $k$NN graphs or $\epsilon$-graphs).

\textbf{Heterogeneous GSL.} We integrate two representative heterogeneous GSL algorithms: Graph Transformer Networks (GTN)\cite{yun2019graph} and Heterogeneous Graph Structure Learning (HGSL)\cite{zhao2021heterogeneous}, which can handle the heterogeneity and capture complex interactions in heterogeneous graphs.

\textbf{Graph-level GSL.} 
Graph-level GSL algorithms aim to refine each graph structure in datasets. We select two graph-level algorithms: Hierarchical Graph Pooling with Structure Learning (HGP-SL)\cite{zhang2019hierarchical} and Variational Information Bottleneck guided Graph Structure Learning (VIB-GSL)\cite{sun2022graph}.

\subsection{Benchmark Datasets}
\label{sec:datasets}
To comprehensively and effectively evaluate the characteristics of GSL in the field of graph learning, we have integrated a large number of datasets from various domains for different types of tasks. For node-level tasks, to evaluate the most mainstream task of GSL, node classification, we use four citation networks (i.e., Cora, Citeseer, Pubmed\cite{yang2016revisiting}), and ogbn-arxiv\cite{hu2020open}, three website networks from WebKB (i.e., Cornell, Texas, and Wisconsin\cite{pei2020geom}), and a cooccurrence network Actor with homophily ratio ranging from strong homophily to strong heterophily. Subsequently, to validate the effectiveness of GSL in heterogeneous node classification, we utilized three heterogeneous graph datasets (i.e., DBLP\cite{yun2019graph}, ACM\cite{yun2019graph}, and Yelp\cite{lu2019relation}). To investigate the robustness of GSL, we further incorporate the Polblogs dataset for evaluation. For graph-level tasks, we select six public graph classification benchmark dataset from TUDataset\cite{morris2020tudataset} for evaluation, including IMDB-B\cite{cai2018simple}, IMDB-M\cite{cai2018simple}, RDT-B\cite{yanardag2015deep}, COLLAB\cite{yanardag2015deep}, MUTAG\cite{debnath1991structure} and PROTEINS\cite{borgwardt2005protein}. Each dataset is a collection of graphs where each graph is associated with a level. Besides, exploring whether GSL can capture long-range information is an exciting topic. Therefore, we have utilized recently proposed long-range graph datasets: Peptides-func and Peptides-struct\cite{dwivedi2022long}. See more details and statistics about datasets in \appref{appen:dataset}.

\subsection{Benchmark Evaluations}
\label{sec:evaluations}
To comprehensively investigate the pros and cons of GSL, our benchmark evaluations encompass three dimensions: \myblue{effectiveness}, \myorange{robustness}, and \mygreen{complexity}. For \myblue{effectiveness}, GSLB provides a fair and comprehensive comparison of existing algorithms from three perspectives: homogeneous node classification, heterogeneous node classification, and graph-level tasks. In the case of homogeneous node classification, we evaluated them on both homophilic and heterophilic graph datasets, conducting experiments in both TR and TI scenarios. For graph-level, we evaluate graph-level GSL algorithms on TUDataset and long-range graph datasets for exploring the capabilities on graph-level tasks. For most datasets, we use accuracy as our evaluation metric. For \myorange{robustness}, GSLB evaluates three types of robustness: supervision signal robustness, structure robustness, and feature robustness. We control the count of labels to explore the supervision signal robustness of GSL and find that GSL exhibits excellent performance in the scenarios with few labels. We inject random structure noise and graph topology attacks to investigate the structure robustness. We also study the feature robustness by randomly masking a certain proportion of node features. For \mygreen{complexity}, we conduct a detailed evaluation of representative GSL algorithms in terms of time complexity and space complexity. It will help to facilitate the deployment of GSL in real-world applications.

\section{Experiments and Analysis}

In this section, we systematically investigate the \myblue{effectiveness}, \myorange{robustness}, and \mygreen{complexity} of GSL algorithms by answering the following specific questions:
\begin{itemize}
    \item For \myblue{effectiveness}, \textbf{RQ1}: How effective are the algorithms on node-level representation learning (\secref{sec:exp-node})? \textbf{RQ2}: Can GSL mitigate homophily inductive bias of traditional message-passing based GNNs (\secref{sec:exp-node})? \textbf{RQ3}: How does GSL perform on heterogeneous graph datasets (\secref{sec:hetero})? \textbf{RQ4}: How effective are the algorithms on graph-level representation learning (\secref{sec:graph})? \textbf{RQ5}: Can GSL methods capture long-range information on the graph (\appref{appen:exp})?
    \item For \myorange{robustness}, \textbf{RQ6}: How robust are GSL algorithms when faced with a scarcity of labeled samples? \textbf{RQ7}: How robust are GSL algorithms in the face of structure attack or noise? \textbf{RQ8}: How is the feature robustness of GSL? (\secref{sec:robust})
    \item For \mygreen{complexity}, \textbf{RQ9}: How efficient are these algorithms in terms of time and space (\secref{sec:complexity})?
    \item Otherwise, \textbf{RQ10}: What does the learned graph structure look like (\appref{appen:visual})?
\end{itemize}

\subsection{Experimental Settings}
\label{sec:settings}
All algorithms in GSLB are implemented by PyTorch \cite{paszke2019pytorch}, and unless specifically indicated, the encoders for all algorithms are Graph Convolutional Networks. All experiments are conducted on a Linux server with GPU (NVIDIA GeForce 3090 and NVIDIA A100) and CPU (AMD EPYC 7763), using PyTorch 1.13.0, DGL 1.1.0\cite{wang2019deep} and Python 3.9.16.

\subsection{Performance on node-level representation learning}
\label{sec:exp-node}

For node-level representation learning, we conduct experiments on homogeneous graph datasets under both TR and TI scenarios, and use classification accuracy as our evaluation metric. \tabref{tab:node-tr} shows the experimental results of various GSL algorithms under the standard setting of transductive node classification task in the TR scenario. We can observe that: 1) Most GSL algorithms generally show improvements in node classification task, particularly on datasets with high heterophily ratio. Due to the presence of heterophilic connections in heterophily graphs, where nodes are often connected to nodes with different labels, it violates the homophily assumption of message-passing neural networks. As a result, traditional GNNs like GCN and GAT exhibit poor performance. However, GSL can improve significantly on heterophily graph datasets by learning new graph structures based on downstream tasks and specific learning objectives, thus enhancing the homophily of the graph and promoting the performance on node-level representation learning. 2) SUBLIME achieves optimal or near-optimal results on most datasets. It learns graph structure through contrastive learning in an unsupervised manner. As mentioned in the recent literature \cite{fatemi2021slaps}, optimizing graph structures solely based on label information is insufficient. Leveraging a large and abundant amount of unlabeled information can enhance the performance of GSL. 3) The scalability of GSL still needs improvement, as only a few models can be trained on large-scale datasets (e.g., ogbn-arxiv). We will discuss the scalability of GSL algorithms in detail in a subsequent section (\secref{sec:complexity}).

\tabref{tab:node-ti} shows the experimental results of the transductive node classification task in the TI scenario. Some GSL algorithms are designed for TR scenario (i.e., GRCN, IDGL, etc.), so we use kNN graphs as their original graph structure. As we can observe, on the homophily graph datasets, GSL outperforms baselines, such as MLP, GCN$_{knn}$ and GAT$_{knn}$. However, on the heterophily graph datasets, most GSL algorithms often have difficulty achieving better results than baseline models. As mentioned in earlier literature, a network with randomness tends to get better performance utilizing kNN for direct information propagation\cite{jin2021universal}. Therefore, traditional message-passing neural networks with kNN graphs demonstrate powerful performance. In addition, as observed in the TR scenario, models that leverage self-supervision to extract abundant unlabeled information often achieve better performance.

\begin{table}[]
\renewcommand\arraystretch{1.1}
\caption{Accuracy $\pm$ STD comparison (\%) under the standard setting of transductive node classification task in the Topology Refinement (TR) scenario, which means the original graph structure is available for each method. Performance is averaged from 10 independent repetitions. The highest results are highlighted with \colorbox[HTML]{B8E7E1}{\textbf{bold}}, while the second highest results are marked with \colorbox[HTML]{D4FAFC}{\underline{underline}}. "OOM" denotes out of memory.}
\label{tab:node-tr}
\resizebox{1\textwidth}{!}{
\begin{tabular}{lcccccccc}
\toprule
       & Cora & Citeseer & Pubmed & ogbn-arxiv & Cornell & Texas & Wisconsin & Actor \\ 
Edge Hom. & 0.81 & 0.74 & 0.80 & 0.65 & 0.12 & 0.06 & 0.18 & 0.22 \\ \midrule
GCN          &  81.46{\tiny$\pm$0.58}  &  71.36{\tiny$\pm$0.31}  &  79.18{\tiny$\pm$0.29}  & \second\underline{ 70.77{\tiny$\pm$0.19}}  &  47.84{\tiny$\pm$5.55}  &  57.83{\tiny$\pm$2.76}  &  57.45{\tiny$\pm$4.30}  &  30.01{\tiny$\pm$0.77}         \\
GAT          &  81.41{\tiny$\pm$0.77}  &  70.69{\tiny$\pm$0.58}   &  77.85{\tiny$\pm$0.42}  &  69.90{\tiny$\pm$0.25}  &  46.22{\tiny$\pm$6.33}  &  54.05{\tiny$\pm$7.35}  &  57.65{\tiny$\pm$7.75}  &  28.91{\tiny$\pm$0.83}      \\ 
GPRGNN       & 83.66{\tiny$\pm$0.77}  & 71.64{\tiny$\pm$0.49}  & 75.99{\tiny$\pm$1.63}  & 50.80{\tiny$\pm$0.29}  & \best\textbf{76.76{\tiny$\pm$5.30}}  & \best\textbf{85.14{\tiny$\pm$3.68}}  & \best\textbf{83.33{\tiny$\pm$3.42}}  & \best\textbf{34.09{\tiny$\pm$1.09}}\\
LDS          & 83.01{\tiny$\pm$0.41}  & \best\textbf{ 73.55{\tiny$\pm$0.54}}  & OOM  & OOM  &  47.87{\tiny$\pm$7.14}  &  58.92{\tiny$\pm$4.32}  &  61.70{\tiny$\pm$3.58} &  31.05{\tiny$\pm$1.31}       \\
GRCN         &  83.87{\tiny$\pm$0.49}  &  72.43{\tiny$\pm$0.61}  &  78.92{\tiny$\pm$0.39}  & OOM          &  54.32{\tiny$\pm$8.24}  & 62.16{\tiny$\pm$7.05}  &  56.08{\tiny$\pm$7.19}  &  29.97{\tiny$\pm$0.71}          \\
ProGNN     &   80.30{\tiny$\pm$0.57} &   68.51{\tiny$\pm$0.52} & OOM  &            OOM &  54.05{\tiny$\pm$6.16} &  48.37{\tiny$\pm$12.17} &  62.54{\tiny$\pm$7.56} &        22.35{\tiny$\pm$0.88}       \\
IDGL         & \best\textbf{ 83.88{\tiny$\pm$0.42}}  &  72.20{\tiny$\pm$1.18}  & \second\underline{ 80.00{\tiny$\pm$0.38}}  & OOM  &  50.00{\tiny$\pm$8.98}  &  62.43{\tiny$\pm$6.09}  &  59.41{\tiny$\pm$4.11}  &  28.16{\tiny$\pm$1.41}                \\
GEN          &  80.21{\tiny$\pm$1.72}  &  71.15{\tiny$\pm$1.81}  &  78.91{\tiny$\pm$0.69}    & OOM  &      57.02{\tiny$\pm$7.19}    & 65.94{\tiny$\pm$1.38} & 66.07{\tiny$\pm$3.72} &  27.21{\tiny$\pm$2.05}           \\
CoGSL        &  81.76{\tiny$\pm$0.24}  & \second\underline{ 73.09{\tiny$\pm$0.42}}  & OOM            & OOM  &  52.16{\tiny$\pm$3.21}  &  59.46{\tiny$\pm$4.36}  &  58.82{\tiny$\pm$1.52}  & 32.95{\tiny$\pm$1.20}              \\
SUBLIME      &  \second\underline{83.40{\tiny$\pm$0.42}}  &  72.30{\tiny$\pm$1.09}  & \best\textbf{ 80.90{\tiny$\pm$0.94}}      & \best\textbf{ 71.75{\tiny$\pm$0.36}}     & \second\underline{ 70.54{\tiny$\pm$5.98}}  & \second\underline{ 77.03{\tiny$\pm$4.23}}  & \second\underline{ 78.82{\tiny$\pm$6.55}}  & \second\underline{ 33.57{\tiny$\pm$0.68}}               \\
STABLE       &  80.20{\tiny$\pm$0.68}  &  68.91{\tiny$\pm$1.01}  & OOM  & OOM  &  44.03{\tiny$\pm$4.05}  &  55.24{\tiny$\pm$6.04}  &  53.00{\tiny$\pm$5.27}  &  30.18{\tiny$\pm$1.00}    \\
NodeFormer   &  80.28{\tiny$\pm$0.82}  &  71.31{\tiny$\pm$0.98}  &  78.21{\tiny$\pm$1.43}              &  55.40{\tiny$\pm$0.23}  &  42.70{\tiny$\pm$5.51}  &  58.92{\tiny$\pm$4.32}  &  48.43{\tiny$\pm$7.02}  &  25.51{\tiny$\pm$1.77}               \\ 
GSR & 82.48{\tiny$\pm$0.43} & 71.10{\tiny$\pm$0.25}& 78.09{\tiny$\pm$0.53}& OOM & 44.32{\tiny$\pm$2.16} & 60.81{\tiny$\pm$4.87}& 56.86{\tiny$\pm$1.24}& 30.23{\tiny$\pm$0.38}\\
\bottomrule
\end{tabular}}
\end{table}

\begin{table}[]
\renewcommand\arraystretch{1.1}
\caption{Accuracy $\pm$ STD comparison (\%) under the standard setting of transductive node classification task in the Topology Inference (TI) scenario, which means the original graph structure is not available for each method.}
\label{tab:node-ti}
\resizebox{1\textwidth}{!}{
\begin{tabular}{lcccccccc}
\toprule
       & Cora & Citeseer & Pubmed & ogbn-arxiv & Cornell & Texas & Wisconsin & Actor \\ 
Edge Hom. & 0.81 & 0.74 & 0.80 & 0.65 & 0.12 & 0.06 & 0.18 & 0.22 \\ \midrule
MLP          &  58.55{\tiny$\pm$0.80}  &  59.52{\tiny$\pm$0.64}   &  73.00{\tiny$\pm$0.30}  &  55.21{\tiny$\pm$0.11}  &  71.35{\tiny$\pm$6.19}  &  \second\underline{80.27{\tiny$\pm$5.93}}  & \best\textbf{ 84.71{\tiny$\pm$3.14}}  &  35.49{\tiny$\pm$1.04}           \\
GCN$_{knn}$          &  66.10{\tiny$\pm$0.44}  &  68.33{\tiny$\pm$0.89}  &  69.23{\tiny$\pm$0.49}  &  55.21{\tiny$\pm$0.22}  &  \second\underline{75.14{\tiny$\pm$2.65}}  &  75.95{\tiny$\pm$4.43}  &  \second\underline{84.12{\tiny$\pm$3.97}}  &  32.98{\tiny$\pm$0.49}            \\
GAT$_{knn}$          &  64.62{\tiny$\pm$1.04}  &  68.05{\tiny$\pm$1.12}   &  68.76{\tiny$\pm$0.80}  & \second\underline{ 55.92{\tiny$\pm$0.30}}  &  74.05{\tiny$\pm$5.16}  &  76.49{\tiny$\pm$4.99}  &  82.16{\tiny$\pm$4.06}  &  31.67{\tiny$\pm$1.19}                     \\ 
GPRGNN$_{knn}$          &  69.27{\tiny$\pm$0.62}  &  70.29{\tiny$\pm$0.54}   &  68.19{\tiny$\pm$1.19}  & 51.39{\tiny$\pm$0.13}  &  \best\textbf{75.68{\tiny$\pm$2.70}}  &  \best\textbf{81.08{\tiny$\pm$4.18}}  &  \second\underline{84.12{\tiny$\pm$3.22}}  &  34.71{\tiny$\pm$1.51}                     \\
LDS          &  69.87{\tiny$\pm$0.41}  &  \second\underline{72.43{\tiny$\pm$0.61}}  & OOM  & OOM  &  72.65{\tiny$\pm$3.86}  &  70.20{\tiny$\pm$5.07}  &  78.14{\tiny$\pm$4.50}  &  32.39{\tiny$\pm$0.79}          \\
GRCN$_{knn}$         &  69.48{\tiny$\pm$0.66}  &  68.41{\tiny$\pm$0.50}  &  68.96{\tiny$\pm$0.85}               & OOM  &  71.08{\tiny$\pm$6.84}  &  74.32{\tiny$\pm$5.02}  &  78.63{\tiny$\pm$4.92} &  30.83{\tiny$\pm$0.76}             \\
ProGNN$_{knn}$       &  67.11{\tiny$\pm$0.56}  &  64.55{\tiny$\pm$0.95} &  OOM &  OOM &  71.35{\tiny$\pm$4.04} &    71.89{\tiny$\pm$5.69} &   72.94{\tiny$\pm$7.93} &   31.56{\tiny$\pm$1.14}                \\
IDGL$_{knn}$         &  69.74{\tiny$\pm$0.57}  &   66.33{\tiny$\pm$0.84} &   74.01{\tiny$\pm$0.64} & OOM &  72.70{\tiny$\pm$4.75} &  75.40{\tiny$\pm$4.75} &  79.21{\tiny$\pm$3.94} &  33.07{\tiny$\pm$1.37}                       \\
GEN$_{knn}$          &    66.95{\tiny$\pm$1.40}   &     67.29{\tiny$\pm$1.17}  &    69.76{\tiny$\pm$1.53}  & OOM &  71.08{\tiny$\pm$5.54} &    74.59{\tiny$\pm$3.46} &  81.76{\tiny$\pm$2.91} &  31.28{\tiny$\pm$1.06}             \\
CoGSL$_{knn}$        & 66.65{\tiny$\pm$0.37}  & 68.72{\tiny$\pm$0.84}  & OOM  & OOM   & 70.27{\tiny$\pm$3.42}  & 72.70{\tiny$\pm$4.26}  & 76.96{\tiny$\pm$5.25} & 34.52{\tiny$\pm$1.56}              \\
GSR$_{knn}$ & 66.28{\tiny$\pm$0.59} & 66.77{\tiny$\pm$0.62} & 68.49{\tiny$\pm$1.49} & OOM & 70.27{\tiny$\pm$3.62} & 74.86{\tiny$\pm$3.63} & 78.62{\tiny$\pm$5.91} & 33.73{\tiny$\pm$1.12} \\
SLAPS        &  \second\underline{72.28{\tiny$\pm$0.97}}  &  70.71{\tiny$\pm$1.13}  & 74.50{\tiny$\pm$1.47}  & 55.19{\tiny$\pm$0.21}  & 74.59{\tiny$\pm$3.67}  &  79.19{\tiny$\pm$4.99}  & 81.96{\tiny$\pm$3.26}  & \best\textbf{ 37.16{\tiny$\pm$0.91}}     \\
SUBLIME      &  72.74{\tiny$\pm$1.91}  &  \best\textbf{72.63{\tiny$\pm$0.60}}  &  \second\underline{75.08{\tiny$\pm$0.55}}  &  55.57{\tiny$\pm$0.18}  &  72.35{\tiny$\pm$3.57}  &  75.51{\tiny$\pm$5.08}  &  82.14{\tiny$\pm$2.62}  &  32.20{\tiny$\pm$1.02}      \\
NodeFormer   & 54.35{\tiny$\pm$5.33}  & 45.90{\tiny$\pm$5.42}  & 59.83{\tiny$\pm$6.50}  & 55.37{\tiny$\pm$0.23}  & 42.70{\tiny$\pm$5.51}  & 58.92{\tiny$\pm$4.32}  & 48.24{\tiny$\pm$6.63}  & 29.24{\tiny$\pm$1.68}              \\
HES-GSL       & \best\textbf{ 73.68{\tiny$\pm$1.04}}  &  70.12{\tiny$\pm$1.11}  & \best\textbf{ 77.08{\tiny$\pm$0.78}}      & \best\textbf{ 56.46{\tiny$\pm$0.27}} &  66.22{\tiny$\pm$6.19}  &  74.05{\tiny$\pm$6.42}  &  79.61{\tiny$\pm$5.28}  & \second \underline{36.73{\tiny$\pm$0.76}}  \\
\bottomrule
\end{tabular}}
\end{table}

\subsection{Performance on heterogeneous graph node-level representation learning}
\label{sec:hetero}
In this section, we evaluate the performance of GSL algorithms on heterogeneous node classification task and use Macro-F1 and Micro-F1 as our evaluation metrics. \tabref{tab:hetero} shows the experimental results on heterogeneous graph datasets. By observing the results, we can find that: 1) Because GTN and HGSL consider both heterogeneity and structure learning, they generally outperform other models on heterogeneous graph datasets. 2) GSL algorithms generally outperform the vanilla GNN models (e.g. GCN and GAT) since they have learned better structures to facilitate message passing. 3) Due to the majority of GSL algorithms not explicitly accounting for heterogeneity, they may exhibit poorer performance on heterogeneous graph datasets. 4) Some datasets (e.g. Yelp) exhibit stronger heterogeneity, and on such datasets, models that consider heterogeneity (e.g. HAN, GTN, and HGSL) perform significantly better.

\begin{table}[]
\caption{Macro-F1 and Micro-F1 $\pm$ STD comparison (\%) under the standard setting of heterogeneous node classification task.}
\label{tab:hetero}
\begin{center}
\resizebox{0.9\textwidth}{!}{
\begin{tabular}{lcccccc}
\toprule
\multirow{2}{*}{Method} & \multicolumn{2}{c}{ACM}       & \multicolumn{2}{c}{DBLP}      & \multicolumn{2}{c}{Yelp}      \\ 
                        & Macro-F1      & Micro-F1      & Macro-F1      & Micro-F1      & Macro-F1      & Micro-F1      \\
\midrule
GCN                     &  90.27{\tiny$\pm$0.59} &  90.18{\tiny$\pm$0.61} &  90.01{\tiny$\pm$0.32} &  90.99{\tiny$\pm$0.28} &  78.01{\tiny$\pm$1.89} &  81.03{\tiny$\pm$1.81} \\
GAT                     &  91.52{\tiny$\pm$0.62} &  91.46{\tiny$\pm$0.62} &  90.22{\tiny$\pm$0.37} &  91.13{\tiny$\pm$0.40} &  82.12{\tiny$\pm$1.47} &  84.43{\tiny$\pm$1.56}             \\
HAN                     &  91.67{\tiny$\pm$0.39} &  91.47{\tiny$\pm$0.22} &  90.53{\tiny$\pm$0.24} &  91.47{\tiny$\pm$0.22} &  88.49{\tiny$\pm$1.73} &  88.78{\tiny$\pm$1.40} \\
LDS                     &  92.35{\tiny$\pm$0.43} &  92.05{\tiny$\pm$0.26} &  88.11{\tiny$\pm$0.86} &  88.74{\tiny$\pm$0.85} &  75.98{\tiny$\pm$2.35} &  78.14{\tiny$\pm$1.98}              \\
GRCN                    &  \second\underline{93.04{\tiny$\pm$0.17}} &  \second\underline{92.94{\tiny$\pm$0.18}} &  88.33{\tiny$\pm$0.47} &  89.43{\tiny$\pm$0.44} &  76.05{\tiny$\pm$1.05} &  80.68{\tiny$\pm$0.96}              \\
IDGL                    &  91.69{\tiny$\pm$1.24} &  91.63{\tiny$\pm$1.24} &  89.65{\tiny$\pm$0.60} &  90.61{\tiny$\pm$0.56} &  76.98{\tiny$\pm$5.78} &  79.15{\tiny$\pm$5.06} \\
ProGNN                  &  90.57{\tiny$\pm$1.03} &  90.50{\tiny$\pm$1.29} &  83.13{\tiny$\pm$1.56} &  84.83{\tiny$\pm$1.36} &  51.76{\tiny$\pm$1.46} &  58.39{\tiny$\pm$1.25}         \\
GEN                  &  87.91{\tiny$\pm$2.78} &  87.88{\tiny$\pm$2.61} &  89.74{\tiny$\pm$0.69} &  90.65{\tiny$\pm$0.71} &  80.43{\tiny$\pm$3.78} &  82.68{\tiny$\pm$2.84}         \\
STABLE                  &  83.54{\tiny$\pm$4.20} &  83.38{\tiny$\pm$4.51} &  75.18{\tiny$\pm$1.95} &  76.42{\tiny$\pm$1.95} &  71.48{\tiny$\pm$4.71} &  76.62{\tiny$\pm$2.75}         \\
GEN                  &  87.91{\tiny$\pm$2.78} &  87.88{\tiny$\pm$2.61} &  89.74{\tiny$\pm$0.69} &  90.65{\tiny$\pm$0.71} &  80.43{\tiny$\pm$3.78} &  82.68{\tiny$\pm$2.84}         \\
SUBLIME                  &  92.42{\tiny$\pm$0.16} &  92.13{\tiny$\pm$0.37} &  \second\underline{90.98{\tiny$\pm$0.37}} &  \second\underline{91.82{\tiny$\pm$0.27}} &  79.68{\tiny$\pm$0.79} &  82.99{\tiny$\pm$0.82}         \\
NodeFormer                  &  91.33{\tiny$\pm$0.77} &  90.60{\tiny$\pm$0.95} &  79.54{\tiny$\pm$0.78} &  80.56{\tiny$\pm$0.62} &  91.69{\tiny$\pm$0.65} &  90.59{\tiny$\pm$1.21}         \\
GSR                  &  92.14{\tiny$\pm$1.08} &  92.11{\tiny$\pm$0.99} &  76.59{\tiny$\pm$0.45} &  77.69{\tiny$\pm$0.42} &  83.85{\tiny$\pm$0.76} &  85.73{\tiny$\pm$0.54}         \\
GTN                     & 92.04{\tiny$\pm$0.38}  & 91.94{\tiny$\pm$0.39}  & 90.52{\tiny$\pm$0.45}  & 91.48{\tiny$\pm$0.39}  & \best\textbf{92.98{\tiny$\pm$0.52}}  & \best\textbf{92.44{\tiny$\pm$0.46}}              \\
HGSL                    & \best\textbf{93.23{\tiny$\pm$0.50}} & \best\textbf{ 93.13{\tiny$\pm$0.51}} & \best\textbf{ 91.58{\tiny$\pm$0.40}} & \best\textbf{ 92.49{\tiny$\pm$0.35}} & \second\underline{ 92.79{\tiny$\pm$0.44}} & \second\underline{ 92.24{\tiny$\pm$0.48}} \\
\bottomrule
\end{tabular}}
\end{center}
\end{table}

\subsection{Performance of GSL algorithms on graph-level tasks}
\label{sec:graph}
In this section, we conduct graph classification experiments on four social datasets (i.e., \texttt{IMDB-B}, \texttt{RDT-B}, \texttt{COLLAB}, and \texttt{IMDB-M}) and two biological datasets (i.e., \texttt{MUTAG} and \texttt{PROTEINS}). \tabref{tab:graph} shows the experimental results of average accuracy and the standard deviation of 10-fold cross-validation. We can observe that HGP-SL (with GCN as the encoder) consistently outperforms GCN on all datasets. However, we find that VIB-GSL exhibits strong instability across different random seeds. And due to the absence of training scripts in the official code\footnote{\url{https://github.com/RingBDStack/VIB-GSL}}, we performed hyperparameter tuning based on the parameter search space ($\beta\in\{10^{-1},10^{-2},10^{-3},10^{-4},10^{-5},10^{-6}\}$) provided in the paper, but we are unable to surpass the performance of the baseline models consistently. Lastly, we conducted an analysis of graph-level GSL algorithms on long-range graph dataset\cite{dwivedi2022long}. For detailed information, please refer to \appref{appen:exp}.

\begin{table}[]
\caption{Accuracy $\pm$ STD comparison (\%) under the setting of graph-level classification task.}
\label{tab:graph}
\begin{center}
\begin{tabular}{lcccccc}
\toprule
Method         & IMDB-B   & RDT-B & COLLAB        & IMDB-M    & MUTAG         & PROTEINS      \\
\midrule
GCN            &  73.20{\tiny$\pm$4.29} &  70.10{\tiny$\pm$5.80}  &  76.96{\tiny$\pm$2.28} &  49.85{\tiny$\pm$3.84} &  73.92{\tiny$\pm$8.84} &  67.52{\tiny$\pm$6.71} \\
VIB-GSL (GCN)  &  71.90{\tiny$\pm$4.48}  &  68.95{\tiny$\pm$2.66}  &  77.14{\tiny$\pm$1.59}  &  49.05{\tiny$\pm$5.52}  &  68.63{\tiny$\pm$5.15}  &  65.68{\tiny$\pm$8.53}              \\
HGP-SL (GCN)         &  \best\textbf{74.10{\tiny$\pm$4.55}}  & OOM  &  78.06{\tiny$\pm$2.17}  &  \best\textbf{51.07{\tiny$\pm$2.00}}  &  78.07{\tiny$\pm$10.85}  &  \second\underline{70.80{\tiny$\pm$4.25}}      \\
\noalign{\vskip 1.5pt}
\hdashline
\noalign{\vskip 1.5pt}
GAT            &  72.30{\tiny$\pm$2.26}  &  \second\underline{73.55{\tiny$\pm$4.76}}  &  79.08{\tiny$\pm$1.36}  &  48.90{\tiny$\pm$2.98}  &  78.71{\tiny$\pm$7.51}  &  68.63{\tiny$\pm$6.24}              \\
VIB-GSL (GAT)  &  72.10{\tiny$\pm$5.69}  & OOM  &  77.54{\tiny$\pm$1.85}  &  49.06{\tiny$\pm$4.55}  &  77.13{\tiny$\pm$9.95}  &  67.09{\tiny$\pm$8.43}             \\
\noalign{\vskip 1.5pt}
\hdashline
\noalign{\vskip 1.5pt}
SAGE           &  72.60{\tiny$\pm$3.69}  &  70.20{\tiny$\pm$4.11}  &  75.58{\tiny$\pm$2.04}  &  48.55{\tiny$\pm$2.03}  &  68.65{\tiny$\pm$4.31}  &  64.47{\tiny$\pm$7.15}              \\
VIB-GSL (SAGE) &  73.00{\tiny$\pm$4.78}  &  65.75{\tiny$\pm$3.17}  &  77.74{\tiny$\pm$1.52}  &  48.79{\tiny$\pm$5.06}  &  72.81{\tiny$\pm$11.41}  &  66.61{\tiny$\pm$4.48}              \\
HGP-SL (SAGE)  &  
 71.50{\tiny$\pm$5.24} &  
OOM &    
 78.64{\tiny$\pm$1.47} &    
 49.67{\tiny$\pm$3.09} &   
 77.13{\tiny$\pm$3.29} & 
 73.32{\tiny$\pm$2.06}  \\
\noalign{\vskip 1.5pt}
\hdashline
\noalign{\vskip 1.5pt}
GIN            &  73.00{\tiny$\pm$2.67}  &  71.70{\tiny$\pm$5.01}  &  \second\underline{79.86{\tiny$\pm$1.64}}  &  \second\underline{50.30{\tiny$\pm$3.52}}  &  \best\textbf{87.19{\tiny$\pm$8.05}}  &   69.07{\tiny$\pm$5.62}              \\
VIB-GSL (GIN)  &  69.90{\tiny$\pm$3.90}  &  \best\textbf{75.85{\tiny$\pm$3.63}}  &  77.25{\tiny$\pm$2.34}  &  49.97{\tiny$\pm$3.65}  &  \second\underline{85.18{\tiny$\pm$10.11}}  &  \best\textbf{75.15{\tiny$\pm$5.72}}              \\
HGP-SL (GIN)  &
\second\underline{73.50{\tiny$\pm$6.25}} &
OOM &
\best\textbf{80.14{\tiny$\pm$1.51}} & 
48.67{\tiny$\pm$2.58}&
73.92{\tiny$\pm$6.24}&
69.37{\tiny$\pm$3.95}\\
\bottomrule
\end{tabular}
\end{center}
\end{table}

\subsection{Robustness analysis of GSL algorithms}
\label{sec:robust}
To investigate the robustness of GSL algorithms, we primarily focus on three aspects: structure robustness, feature robustness, and supervision signal robustness. Due to limited space, we predominantly investigate the transductive node classification task in our paper. Nevertheless, researchers can utilize our GSLB library to efficiently and conveniently conduct experiments on other tasks as well.

\begin{wrapfigure}{r}{0.65\textwidth}
  \centering
  \includegraphics[width=\linewidth]{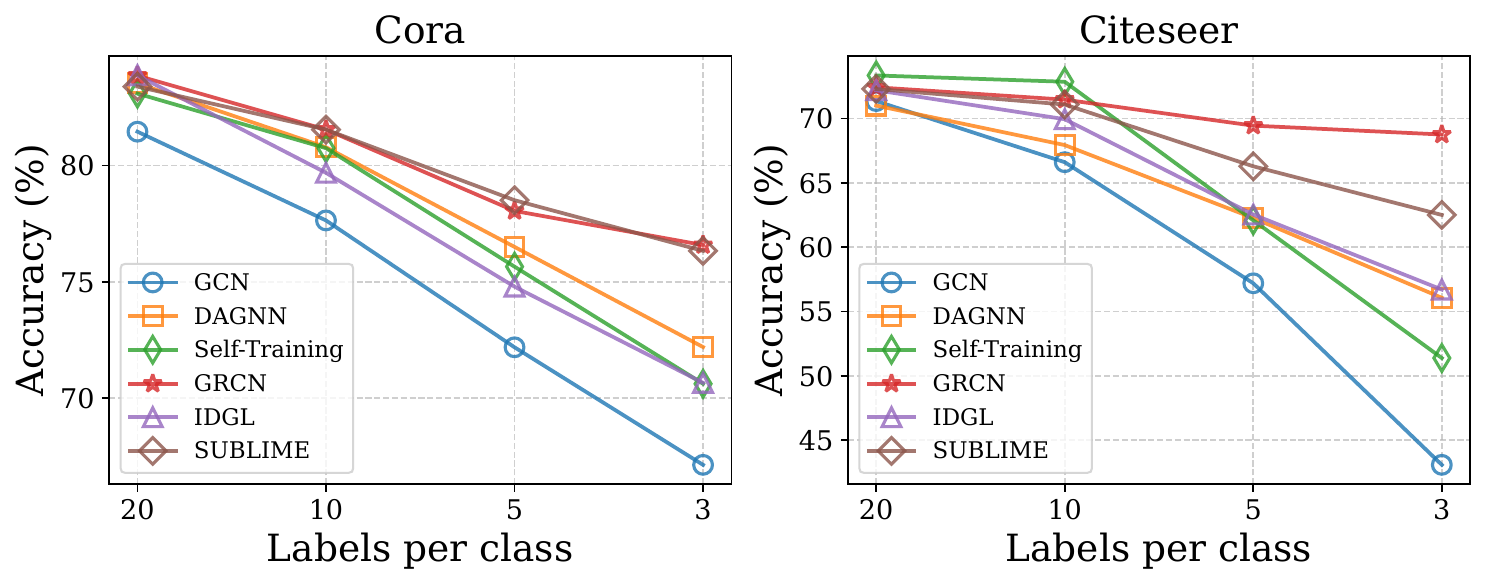}
  \caption{Performance of baselines and GSL algorithms with respect to different numbers of labels per class on Cora and Citeseer.}
  \label{fig:label}
\end{wrapfigure}

\textbf{Robustness analysis with respect to different supervision signals.} We have discovered that GSL maintains surprising performance in scenarios with a scarcity of labeled samples. We varied the number of labels per class in the Cora and Citeseer datasets and selected two baseline models, DAGNN\cite{liu2020towards} and Self-Training\cite{li2018deeper}, that performed well in scenarios with limited labels. As shown in \figref{fig:label}, we can observe that GSL algorithms (for the sake of brevity, we opted to select only three models: GRCN, IDGL, and SUBLIME) achieve the best results in scenarios with fewer labels available. We speculate that this may be because the learned graph structure is denser and exhibits cleaner community boundaries. As a result, the supervision signals can propagate more effectively within such a structure.

\begin{figure}[t]
\centering
\includegraphics[width=\textwidth]{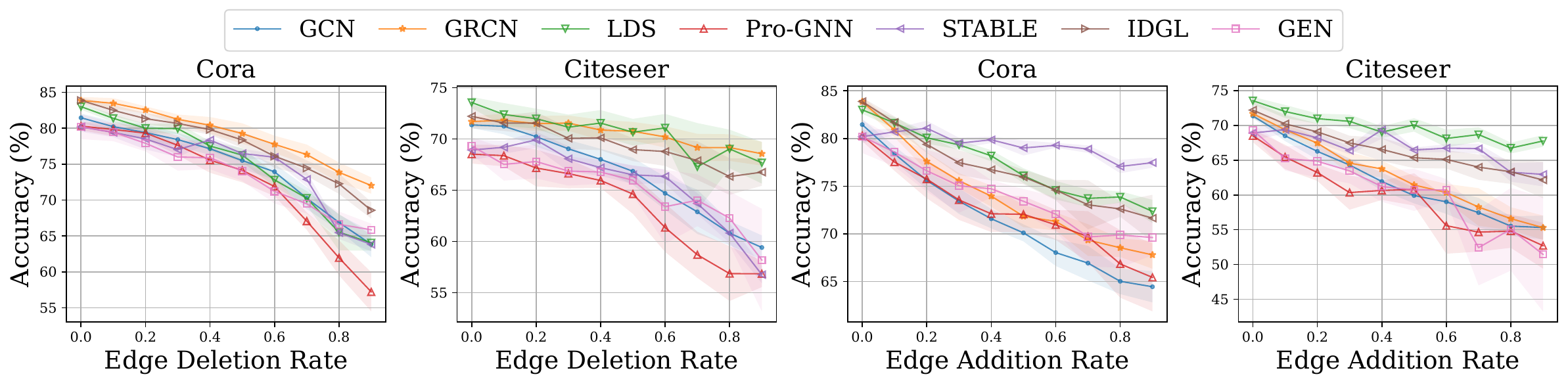}
\caption{Analysis of robustness when injecting random noise on Cora and Citeseer.}
\label{fig:random}
\end{figure}

\textbf{Robustness analysis with respect to random noise.} We randomly remove edges from or add edges to the original graph structures of \texttt{Cora} and \texttt{Citeseer}, then evaluated the performance of GSL algorithms on the corrupted graphs. We change the ratios of modified edges from 0 to 0.9 to simulate different attack intensities. 
As shown in \figref{fig:random}, as the noise intensity increases, the models' performance generally exhibits a downward trend. And we can observe that GSL algorithms commonly demonstrate a certain degree of robustness, as they tend to exhibit more stable performance than GCN when random noise is injected. Besides, we also found that, due to variations in the graph modeling process, different algorithms display varying levels of robustness when facing edge deletion and edge addition scenarios. For example, GRCN demonstrates strong robustness in edge deletion scenarios. However, in the edge addition scenarios, it only exhibits slight performance improvements compared to GCN. On the contrary, STABLE exhibits strong robustness in the edge deletion scenario, while showing the opposite trend in edge addition.

\begin{table}[]
\caption{Accuracy $\pm$ STD comparison (\%) with respect to different perturbation rates. \textcolor[HTML]{820000}{Jaccard} and \textcolor[HTML]{820000}{SimPGCN} are representative state-of-the-art defense GNNs.}
\label{tab:attack}
\begin{tabular}{lccccccccc}
\toprule
Dataset                   & Ptb Rate & GCN & \textcolor[HTML]{820000}{Jaccard} & \textcolor[HTML]{820000}{SimPGCN} & IDGL & GRCN & ProGNN  & STABLE & SUBLIME \\
\midrule
\multirow{5}{*}{Cora}     & 0\%      & 83.68{\tiny$\pm$0.37}    & 83.78{\tiny$\pm$0.50}    & 82.66{\tiny$\pm$0.48}    & \best\textbf{84.69{\tiny$\pm$1.13}}    & 84.43{\tiny$\pm$0.26}   & \second\underline{84.53{\tiny$\pm$0.89}}           & 83.70{\tiny$\pm$0.30}       & 83.84{\tiny$\pm$0.28}        \\
                          & 5\%      & 80.61{\tiny$\pm$0.39}    & 81.44{\tiny$\pm$0.48}    & 80.35{\tiny$\pm$0.82}    & \best\textbf{82.56{\tiny$\pm$0.24}}    & 81.34{\tiny$\pm$0.50}    & 81.47{\tiny$\pm$0.44}           & \second\underline{81.52{\tiny$\pm$0.85}}       & 79.93{\tiny$\pm$0.58}        \\
                          & 10\%     & 74.38{\tiny$\pm$0.59}    & 75.90{\tiny$\pm$0.64}    & 76.50{\tiny$\pm$1.12}    & 78.06{\tiny$\pm$0.62}    & 77.12{\tiny$\pm$0.38}    & 72.61{\tiny$\pm$0.73}           & \second\underline{78.64{\tiny$\pm$1.82}}       & \best\textbf{78.71{\tiny$\pm$0.46}}        \\
                          & 15\%     & 65.17{\tiny$\pm$0.99}    & 77.14{\tiny$\pm$0.70}    & 73.77{\tiny$\pm$1.88}    & 76.88{\tiny$\pm$0.44}    & 73.74{\tiny$\pm$0.61}    & 65.68{\tiny$\pm$1.97}           & \best\textbf{79.70{\tiny$\pm$1.71}}       & \second\underline{79.34{\tiny$\pm$0.61}}        \\
                          & 20\%     & 61.98{\tiny$\pm$1.23}    & 70.71{\tiny$\pm$0.91}    & 69.08{\tiny$\pm$2.78}    & 67.19{\tiny$\pm$0.69}    & 69.54{\tiny$\pm$0.58}   & 61.07{\tiny$\pm$0.61}           & \best\textbf{76.44{\tiny$\pm$2.47}}       & \second\underline{75.25{\tiny$\pm$1.08}}        \\
\midrule
\multirow{5}{*}{Citeseer} & 0\%      & \best\textbf{76.56{\tiny$\pm$0.36}}    & 74.34{\tiny$\pm$0.26}    & 74.35{\tiny$\pm$0.74}    & 73.87{\tiny$\pm$0.70}    & \second\underline{76.34{\tiny$\pm$0.11}}    & 73.36{\tiny$\pm$1.52}          & 72.65{\tiny$\pm$1.36}       & 73.34{\tiny$\pm$1.17}        \\
                          & 5\%      & 72.51{\tiny$\pm$0.30}    & 70.01{\tiny$\pm$0.79}    & 72.99{\tiny$\pm$1.05}    & 72.46{\tiny$\pm$0.47}    & \best\textbf{74.66{\tiny$\pm$0.27}}    & 71.46{\tiny$\pm$0.47}          & 69.66{\tiny$\pm$0.95}       & \second\underline{72.63{\tiny$\pm$0.50}}        \\
                          & 10\%     & 71.92{\tiny$\pm$0.68}    & 70.28{\tiny$\pm$1.30}    & 72.68{\tiny$\pm$0.54}    & 69.72{\tiny$\pm$0.59}    & \best\textbf{74.06{\tiny$\pm$0.43}}    & 69.03{\tiny$\pm$0.60}          & 72.79{\tiny$\pm$0.71}       & \second\underline{73.02{\tiny$\pm$0.29}}        \\
                          & 15\%     & 64.44{\tiny$\pm$0.53}    & 67.13{\tiny$\pm$1.28}    & \second\underline{71.74{\tiny$\pm$1.46}}    & 62.83{\tiny$\pm$1.28}    & 66.46{\tiny$\pm$1.12}    & 65.42{\tiny$\pm$1.20}          & 70.98{\tiny$\pm$0.61}       & \best\textbf{73.90{\tiny$\pm$0.52}}        \\
                          & 20\%     & 57.51{\tiny$\pm$1.03}    & 67.82{\tiny$\pm$0.74}    & 70.06{\tiny$\pm$1.86}    & 61.16{\tiny$\pm$0.99}    & 69.42{\tiny$\pm$1.14}    & 57.51{\tiny$\pm$0.36}          & \second\underline{71.90{\tiny$\pm$1.12}}       & \best\textbf{72.55{\tiny$\pm$0.62}}        \\
\midrule
\multirow{5}{*}{Polblogs}   & 0\%      & 95.62{\tiny$\pm$0.69}    & 94.93{\tiny$\pm$0.28}    & 94.50{\tiny$\pm$0.43}    & 94.83{\tiny$\pm$0.20}    & \best\textbf{95.65{\tiny$\pm$0.28}}  &       94.84{\tiny$\pm$0.19}    & \second\underline{95.63{\tiny$\pm$0.32}}       & 95.27{\tiny$\pm$0.51}        \\
                          & 5\%      & 80.57{\tiny$\pm$0.66}    & 78.17{\tiny$\pm$0.55}    & 76.02{\tiny$\pm$1.14}    & 79.62{\tiny$\pm$0.65}    & \best\textbf{93.70{\tiny$\pm$0.18}}  &       92.36{\tiny$\pm$0.42}    & 89.41{\tiny$\pm$1.63}       & \second\underline{93.24{\tiny$\pm$1.50}}        \\
                          & 10\%     & 71.83{\tiny$\pm$2.37}    & 71.86{\tiny$\pm$1.34}    & 70.12{\tiny$\pm$1.10}    & 74.54{\tiny$\pm$0.69}    & 87.99{\tiny$\pm$1.56}   &     84.66{\tiny$\pm$0.52}      & 89.87{\tiny$\pm$0.82}       & \best\textbf{93.62{\tiny$\pm$0.50}}        \\
                          & 15\%     & 66.38{\tiny$\pm$2.17}    & 69.93{\tiny$\pm$0.66}    & 64.19{\tiny$\pm$1.55}    & 75.53{\tiny$\pm$0.83}    & 71.85{\tiny$\pm$1.58}   &      77.38{\tiny$\pm$0.51}     & 89.94{\tiny$\pm$0.89}       & \best\textbf{94.29{\tiny$\pm$0.27}}        \\
                          & 20\%     & 68.19{\tiny$\pm$2.24}    & 69.22{\tiny$\pm$0.34}    & 63.64{\tiny$\pm$1.41}    & 71.63{\tiny$\pm$0.62}    & 71.73{\tiny$\pm$1.58}  &       73.57{\tiny$\pm$0.29}    & \second\underline{87.42{\tiny$\pm$0.69}}       & \best\textbf{92.60{\tiny$\pm$0.72}}       \\
\bottomrule
\end{tabular}
\end{table}

\textbf{Robust analysis with respect to graph topology attack.} Following \cite{li2022reliable, zheng2020robust}, we conduct robust analysis on three graph datasets, i.e., Cora, Citeseer, and Polblogs. First, we select the largest connected component in the graph, and utilize Mettack\cite{1902.08412}, a non-targeted adversarial topology attack method, to generate perturbed graphs. We select the perturbation rate from 0\% to 20\%. \tabref{tab:attack} shows the performance of GSL algorithms on three datasets with respect to various perturbation rates. Surprisingly, we can observe that most GSL algorithms exhibit strong robustness against graph topology attacks, even better than state-of-the-art defense GNNs (e.g., Jaccard\cite{wu2019adversarial} and SimPGCN\cite{jin2021node}). GSL can effectively remove the newly added adversarial edges, and recover important edges to promote message passing. As mentioned in \citet{li2022reliable}, optimizing graph structures based on either features or supervised signals might not be reliable. We found that self-supervised graph structure modeling methods (e.g., STABLE and SUBLIME) show excellent performance on corrupted graph structure datasets, which means unsupervised representation learning might produce more reliable and high-quality representations to conduct structure modeling.

\begin{figure}[t]
\centering
\includegraphics[width=\textwidth]{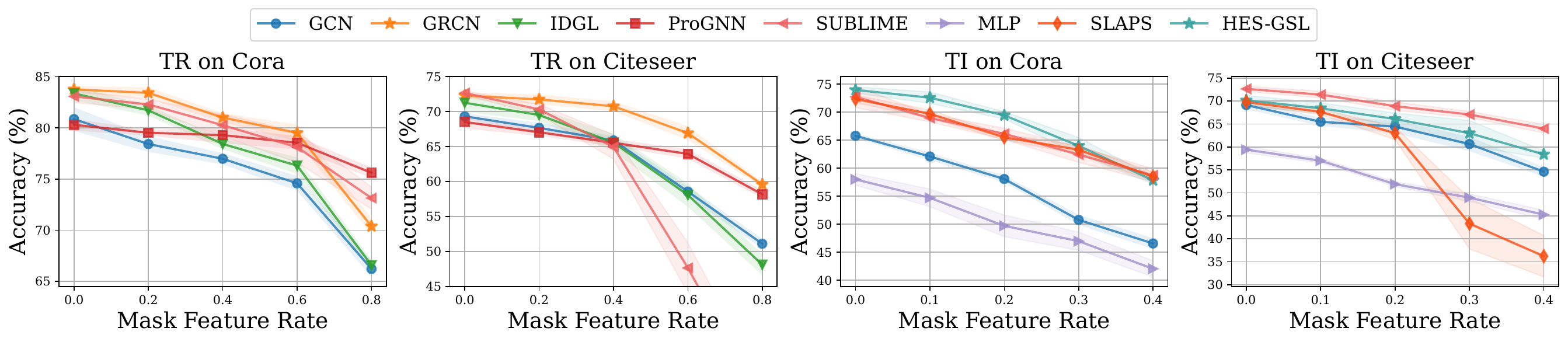}
\caption{Analysis of robustness when injecting random feature noise on Cora and Citeseer.}
\label{fig:feature_robust}
\end{figure}

\textbf{Robust analysis with respect to feature noise.} On the basis of exploring structural robustness, we also study the feature robustness of GSL. We randomly mask a certain proportion of node features by filling them with zeros, to investigate the performance of GSL algorithms when node features are subjected to varying degrees of damage. As shown in \figref{fig:feature_robust}, we can observe that: 1) the node features play a more critical role than the structure on certain datasets. Under the same noise degree, feature noise brings more performance degradation compared with structure noise; 2) Interestingly, while most existing GSL methods rely on feature similarity between pairs of nodes to learn graph structure, they still exhibit good robustness when facing noisy node features; 3) edge-oriented algorithms (e.g., ProGNN) show stronger feature robustness, because they optimize adjacency matrix directly, and have less dependence on pairs of node features.

\begin{figure}[t]
\centering
\includegraphics[width=\textwidth]{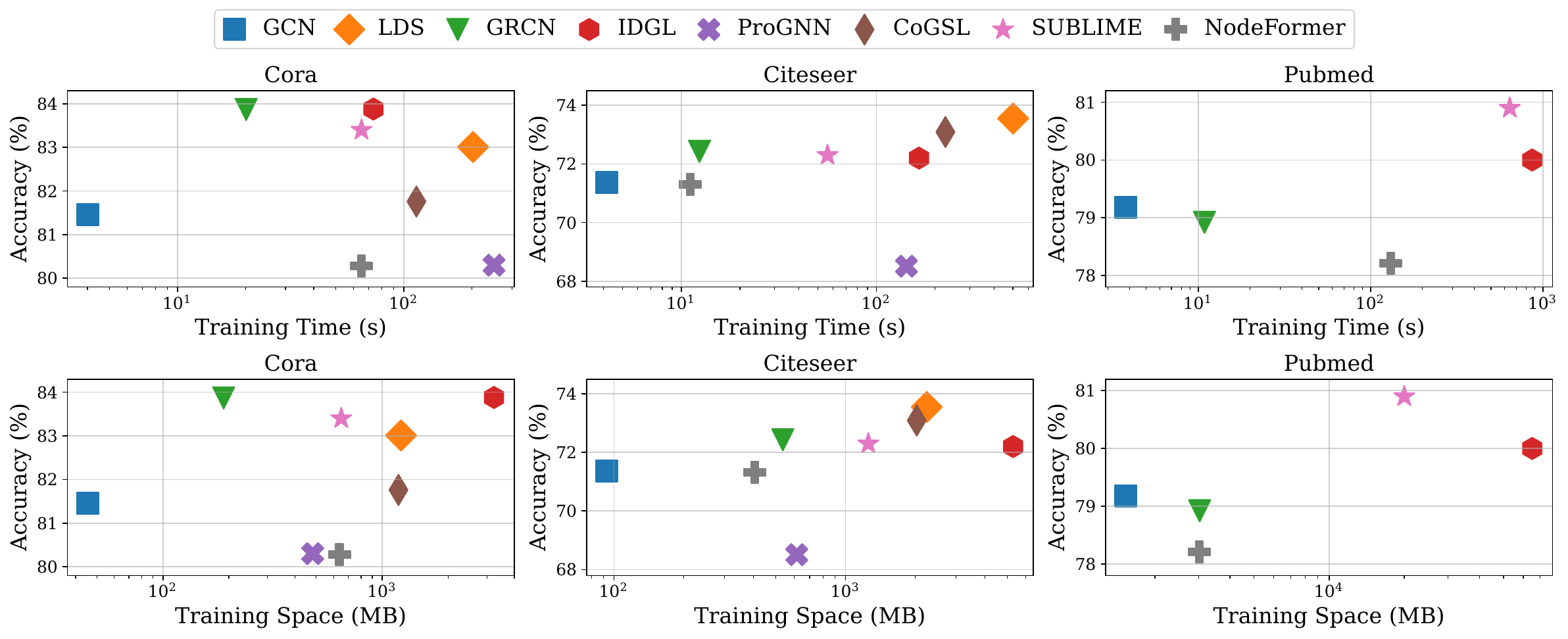}
\caption{Training time and space analysis on \texttt{Cora}, \texttt{Citeseer} and \texttt{Pubmed}.}
\label{fig:complexity}
\end{figure}

\subsection{Efficiency and scalability analysis}
\label{sec:complexity}
In this section, we analyze the efficiency and scalability of GSL algorithms on Cora, Citeseer, and Pubmed datasets. For time efficiency, we evaluate the efficiency of the algorithms by measuring the time it takes for them to converge, i.e., achieve the best performance on the validation set. For scalability, we set all models to their dense version to ensure a fair comparison. As shown in \figref{fig:complexity}, GSL algorithms generally have higher time and space complexity compared to GCN. This limitation restricts the application of GSL on large-scale graphs. We can observe that some algorithms (e.g., GRCN) can achieve relatively good performance improvement with less complexity increase. Besides, although some algorithms (e.g., IDGL, LDS, and SUBLIME) achieve remarkable effectiveness improvement, they largely increase the complexity of time and space.

\section{Conclusion and Future Directions}

In this paper, we give a brief introduction and overview of graph structure learning. Then we present the first Graph Structure Learning Benchmark (GSLB) consisting of 16 algorithms and 20 datasets for various tasks. Based on GSLB, we conducted extensive experiments to reveal and analyze the performance of GSL algorithms in different scenarios and tasks. Through our comparative study, we find that GSL achieves promising results in heterophily, robustness, etc. The goal of this work is to understand the current state of development of GSL and provide insights for future research. 

Notwithstanding the promising results that have been made, there are still some critical challenges and research directions worthy of future investigation.
\begin{itemize}
    \item \textbf{Insufficient scalability}. Most existing works model the existence probability of edges based on node pairs, with a complexity of $O(N^2)$. This makes it challenging to employ GSL in large-scale graphs in real-world applications. Future work should focus on overcoming the limitations of GSL in terms of complexity.
    \item \textbf{Surprising performance with few labels}. We have observed that GSL learns denser and more distinct graph structures, which facilitates the propagation of supervision signals. Most existing GNNs that address few label problem are based on deep GNNs\cite{liu2020towards,gasteiger2018predict} or semi-supervised approaches\cite{ding2022meta,sun2020multi,lee2022grafn}, without refining the graph structure. In the future, it would be worth exploring the combination of increasing the supervision signals and making the graph structure more suitable for propagating those signals.
    \item \textbf{Excellent performance of unsupervised GSL in robustness}. Some algorithms using self-supervised methods for learning graph structures exhibit excellent performance in robustness, which may be attributed to the avoidance of unreliable supervision signals. In the future, further exploration can be done to utilize unsupervised structure learning for designing defense models.
    \item \textbf{Hard to apply on incomplete graphs.} Most existing algorithms rely on pairwise node embeddings to generate the probability of edge existence. The underlying assumption is that all attributes of nodes on the graph are complete. However, it is common in practice that some nodes or all nodes have no features. Future research should address the challenges of structure learning on incomplete graphs.
\end{itemize}

\section*{Acknowledge}

This work was partially supported by the Research Grants Council of Hong Kong, No. 14202919 and No. 14205520.

\bibliographystyle{plainnat}
\bibliography{reference}

\begin{thebibliography}{58}
\providecommand{\natexlab}[1]{#1}
\providecommand{\url}[1]{\texttt{#1}}
\expandafter\ifx\csname urlstyle\endcsname\relax
  \providecommand{\doi}[1]{doi: #1}\else
  \providecommand{\doi}{doi: \begingroup \urlstyle{rm}\Url}\fi

\bibitem[Bian et~al.(2020)Bian, Xiao, Xu, Zhao, Huang, Rong, and
  Huang]{bian2020rumor}
Tian Bian, Xi~Xiao, Tingyang Xu, Peilin Zhao, Wenbing Huang, Yu~Rong, and
  Junzhou Huang.
\newblock Rumor detection on social media with bi-directional graph
  convolutional networks.
\newblock In \emph{Proceedings of the AAAI conference on artificial
  intelligence}, volume~34, pages 549--556, 2020.

\bibitem[Borgwardt et~al.(2005)Borgwardt, Ong, Sch{\"o}nauer, Vishwanathan,
  Smola, and Kriegel]{borgwardt2005protein}
Karsten~M Borgwardt, Cheng~Soon Ong, Stefan Sch{\"o}nauer, SVN Vishwanathan,
  Alex~J Smola, and Hans-Peter Kriegel.
\newblock Protein function prediction via graph kernels.
\newblock \emph{Bioinformatics}, 21\penalty0 (suppl\_1):\penalty0 i47--i56,
  2005.

\bibitem[Cai and Wang(2018)]{cai2018simple}
Chen Cai and Yusu Wang.
\newblock A simple yet effective baseline for non-attributed graph
  classification.
\newblock \emph{arXiv preprint arXiv:1811.03508}, 2018.

\bibitem[Chen et~al.(2020)Chen, Wu, and Zaki]{chen2020iterative}
Yu~Chen, Lingfei Wu, and Mohammed Zaki.
\newblock Iterative deep graph learning for graph neural networks: Better and
  robust node embeddings.
\newblock \emph{Advances in neural information processing systems},
  33:\penalty0 19314--19326, 2020.

\bibitem[Debnath et~al.(1991)Debnath, Lopez~de Compadre, Debnath, Shusterman,
  and Hansch]{debnath1991structure}
Asim~Kumar Debnath, Rosa~L Lopez~de Compadre, Gargi Debnath, Alan~J Shusterman,
  and Corwin Hansch.
\newblock Structure-activity relationship of mutagenic aromatic and
  heteroaromatic nitro compounds. correlation with molecular orbital energies
  and hydrophobicity.
\newblock \emph{Journal of medicinal chemistry}, 34\penalty0 (2):\penalty0
  786--797, 1991.

\bibitem[Ding et~al.(2022)Ding, Wang, Caverlee, and Liu]{ding2022meta}
Kaize Ding, Jianling Wang, James Caverlee, and Huan Liu.
\newblock Meta propagation networks for graph few-shot semi-supervised
  learning.
\newblock In \emph{Proceedings of the AAAI Conference on Artificial
  Intelligence}, volume~36, pages 6524--6531, 2022.

\bibitem[Dobson and Doig(2003)]{dobson2003distinguishing}
Paul~D Dobson and Andrew~J Doig.
\newblock Distinguishing enzyme structures from non-enzymes without alignments.
\newblock \emph{Journal of molecular biology}, 330\penalty0 (4):\penalty0
  771--783, 2003.

\bibitem[Dwivedi et~al.(2022)Dwivedi, Ramp{\'a}{\v{s}}ek, Galkin, Parviz, Wolf,
  Luu, and Beaini]{dwivedi2022long}
Vijay~Prakash Dwivedi, Ladislav Ramp{\'a}{\v{s}}ek, Michael Galkin, Ali Parviz,
  Guy Wolf, Anh~Tuan Luu, and Dominique Beaini.
\newblock Long range graph benchmark.
\newblock \emph{Advances in Neural Information Processing Systems},
  35:\penalty0 22326--22340, 2022.

\bibitem[Fan et~al.(2019)Fan, Ma, Li, He, Zhao, Tang, and Yin]{fan2019graph}
Wenqi Fan, Yao Ma, Qing Li, Yuan He, Eric Zhao, Jiliang Tang, and Dawei Yin.
\newblock Graph neural networks for social recommendation.
\newblock In \emph{The world wide web conference}, pages 417--426, 2019.

\bibitem[Fatemi et~al.(2021)Fatemi, El~Asri, and Kazemi]{fatemi2021slaps}
Bahare Fatemi, Layla El~Asri, and Seyed~Mehran Kazemi.
\newblock Slaps: Self-supervision improves structure learning for graph neural
  networks.
\newblock \emph{Advances in Neural Information Processing Systems},
  34:\penalty0 22667--22681, 2021.

\bibitem[Franceschi et~al.(2019)Franceschi, Niepert, Pontil, and
  He]{franceschi2019learning}
Luca Franceschi, Mathias Niepert, Massimiliano Pontil, and Xiao He.
\newblock Learning discrete structures for graph neural networks.
\newblock In \emph{International conference on machine learning}, pages
  1972--1982. PMLR, 2019.

\bibitem[Gao et~al.(2020)Gao, Hu, and Guo]{gao2020exploring}
Xiang Gao, Wei Hu, and Zongming Guo.
\newblock Exploring structure-adaptive graph learning for robust
  semi-supervised classification.
\newblock In \emph{2020 ieee international conference on multimedia and expo
  (icme)}, pages 1--6. IEEE, 2020.

\bibitem[Gasteiger et~al.(2018)Gasteiger, Bojchevski, and
  G{\"u}nnemann]{gasteiger2018predict}
Johannes Gasteiger, Aleksandar Bojchevski, and Stephan G{\"u}nnemann.
\newblock Predict then propagate: Graph neural networks meet personalized
  pagerank.
\newblock \emph{arXiv preprint arXiv:1810.05997}, 2018.

\bibitem[Hao et~al.(2020)Hao, Lu, Huang, Wang, Hu, Liu, Chen, and
  Lee]{hao2020asgn}
Zhongkai Hao, Chengqiang Lu, Zhenya Huang, Hao Wang, Zheyuan Hu, Qi~Liu, Enhong
  Chen, and Cheekong Lee.
\newblock Asgn: An active semi-supervised graph neural network for molecular
  property prediction.
\newblock In \emph{Proceedings of the 26th ACM SIGKDD International Conference
  on Knowledge Discovery \& Data Mining}, pages 731--752, 2020.

\bibitem[Hu et~al.(2020)Hu, Fey, Zitnik, Dong, Ren, Liu, Catasta, and
  Leskovec]{hu2020open}
Weihua Hu, Matthias Fey, Marinka Zitnik, Yuxiao Dong, Hongyu Ren, Bowen Liu,
  Michele Catasta, and Jure Leskovec.
\newblock Open graph benchmark: Datasets for machine learning on graphs.
\newblock \emph{Advances in neural information processing systems},
  33:\penalty0 22118--22133, 2020.

\bibitem[Jiang et~al.(2019)Jiang, Zhang, Lin, Tang, and Luo]{jiang2019semi}
Bo~Jiang, Ziyan Zhang, Doudou Lin, Jin Tang, and Bin Luo.
\newblock Semi-supervised learning with graph learning-convolutional networks.
\newblock In \emph{Proceedings of the IEEE/CVF conference on computer vision
  and pattern recognition}, pages 11313--11320, 2019.

\bibitem[Jin et~al.(2021{\natexlab{a}})Jin, Yu, Huo, Wang, Wang, He, and
  Han]{jin2021universal}
Di~Jin, Zhizhi Yu, Cuiying Huo, Rui Wang, Xiao Wang, Dongxiao He, and Jiawei
  Han.
\newblock Universal graph convolutional networks.
\newblock \emph{Advances in Neural Information Processing Systems},
  34:\penalty0 10654--10664, 2021{\natexlab{a}}.

\bibitem[Jin et~al.(2020)Jin, Ma, Liu, Tang, Wang, and Tang]{jin2020graph}
Wei Jin, Yao Ma, Xiaorui Liu, Xianfeng Tang, Suhang Wang, and Jiliang Tang.
\newblock Graph structure learning for robust graph neural networks.
\newblock In \emph{Proceedings of the 26th ACM SIGKDD international conference
  on knowledge discovery \& data mining}, pages 66--74, 2020.

\bibitem[Jin et~al.(2021{\natexlab{b}})Jin, Derr, Wang, Ma, Liu, and
  Tang]{jin2021node}
Wei Jin, Tyler Derr, Yiqi Wang, Yao Ma, Zitao Liu, and Jiliang Tang.
\newblock Node similarity preserving graph convolutional networks.
\newblock In \emph{Proceedings of the 14th ACM international conference on web
  search and data mining}, pages 148--156, 2021{\natexlab{b}}.

\bibitem[Lee et~al.(2022)Lee, Oh, In, Lee, Hyun, and Park]{lee2022grafn}
Junseok Lee, Yunhak Oh, Yeonjun In, Namkyeong Lee, Dongmin Hyun, and Chanyoung
  Park.
\newblock Grafn: Semi-supervised node classification on graph with few labels
  via non-parametric distribution assignment.
\newblock In \emph{Proceedings of the 45th International ACM SIGIR Conference
  on Research and Development in Information Retrieval}, pages 2243--2248,
  2022.

\bibitem[Li et~al.(2022{\natexlab{a}})Li, Liu, Ao, Chi, Feng, Yang, and
  He]{li2022reliable}
Kuan Li, Yang Liu, Xiang Ao, Jianfeng Chi, Jinghua Feng, Hao Yang, and Qing He.
\newblock Reliable representations make a stronger defender: Unsupervised
  structure refinement for robust gnn.
\newblock In \emph{Proceedings of the 28th ACM SIGKDD Conference on Knowledge
  Discovery and Data Mining}, pages 925--935, 2022{\natexlab{a}}.

\bibitem[Li et~al.(2018)Li, Han, and Wu]{li2018deeper}
Qimai Li, Zhichao Han, and Xiao-Ming Wu.
\newblock Deeper insights into graph convolutional networks for semi-supervised
  learning.
\newblock In \emph{Proceedings of the AAAI conference on artificial
  intelligence}, volume~32, 2018.

\bibitem[Li et~al.(2022{\natexlab{b}})Li, Chen, Liu, and Wu]{li2022devil}
Zhixun Li, Dingshuo Chen, Qiang Liu, and Shu Wu.
\newblock The devil is in the conflict: Disentangled information graph neural
  networks for fraud detection.
\newblock \emph{arXiv preprint arXiv:2210.12384}, 2022{\natexlab{b}}.

\bibitem[Lim et~al.(2021)Lim, Hohne, Li, Huang, Gupta, Bhalerao, and
  Lim]{lim2021large}
Derek Lim, Felix Hohne, Xiuyu Li, Sijia~Linda Huang, Vaishnavi Gupta, Omkar
  Bhalerao, and Ser~Nam Lim.
\newblock Large scale learning on non-homophilous graphs: New benchmarks and
  strong simple methods.
\newblock \emph{Advances in Neural Information Processing Systems},
  34:\penalty0 20887--20902, 2021.

\bibitem[Liu et~al.(2020)Liu, Gao, and Ji]{liu2020towards}
Meng Liu, Hongyang Gao, and Shuiwang Ji.
\newblock Towards deeper graph neural networks.
\newblock In \emph{Proceedings of the 26th ACM SIGKDD international conference
  on knowledge discovery \& data mining}, pages 338--348, 2020.

\bibitem[Liu et~al.(2022{\natexlab{a}})Liu, Wang, Wu, Chen, Guo, and
  Shi]{liu2022compact}
Nian Liu, Xiao Wang, Lingfei Wu, Yu~Chen, Xiaojie Guo, and Chuan Shi.
\newblock Compact graph structure learning via mutual information compression.
\newblock In \emph{Proceedings of the ACM Web Conference 2022}, pages
  1601--1610, 2022{\natexlab{a}}.

\bibitem[Liu et~al.(2021)Liu, Ao, Qin, Chi, Feng, Yang, and He]{liu2021pick}
Yang Liu, Xiang Ao, Zidi Qin, Jianfeng Chi, Jinghua Feng, Hao Yang, and Qing
  He.
\newblock Pick and choose: a gnn-based imbalanced learning approach for fraud
  detection.
\newblock In \emph{Proceedings of the Web Conference 2021}, pages 3168--3177,
  2021.

\bibitem[Liu et~al.(2022{\natexlab{b}})Liu, Zheng, Zhang, Chen, Peng, and
  Pan]{liu2022towards}
Yixin Liu, Yu~Zheng, Daokun Zhang, Hongxu Chen, Hao Peng, and Shirui Pan.
\newblock Towards unsupervised deep graph structure learning.
\newblock In \emph{Proceedings of the ACM Web Conference 2022}, pages
  1392--1403, 2022{\natexlab{b}}.

\bibitem[Lu et~al.(2019)Lu, Shi, Hu, and Liu]{lu2019relation}
Yuanfu Lu, Chuan Shi, Linmei Hu, and Zhiyuan Liu.
\newblock Relation structure-aware heterogeneous information network embedding.
\newblock In \emph{Proceedings of the AAAI Conference on Artificial
  Intelligence}, volume~33, pages 4456--4463, 2019.

\bibitem[Luo et~al.(2021)Luo, Cheng, Yu, Zong, Ni, Chen, and
  Zhang]{luo2021learning}
Dongsheng Luo, Wei Cheng, Wenchao Yu, Bo~Zong, Jingchao Ni, Haifeng Chen, and
  Xiang Zhang.
\newblock Learning to drop: Robust graph neural network via topological
  denoising.
\newblock In \emph{Proceedings of the 14th ACM international conference on web
  search and data mining}, pages 779--787, 2021.

\bibitem[Morris et~al.(2020)Morris, Kriege, Bause, Kersting, Mutzel, and
  Neumann]{morris2020tudataset}
Christopher Morris, Nils~M Kriege, Franka Bause, Kristian Kersting, Petra
  Mutzel, and Marion Neumann.
\newblock Tudataset: A collection of benchmark datasets for learning with
  graphs.
\newblock \emph{arXiv preprint arXiv:2007.08663}, 2020.

\bibitem[Newman(2003)]{newman2003mixing}
Mark~EJ Newman.
\newblock Mixing patterns in networks.
\newblock \emph{Physical review E}, 67\penalty0 (2):\penalty0 026126, 2003.

\bibitem[Paszke et~al.(2019)Paszke, Gross, Massa, Lerer, Bradbury, Chanan,
  Killeen, Lin, Gimelshein, Antiga, et~al.]{paszke2019pytorch}
Adam Paszke, Sam Gross, Francisco Massa, Adam Lerer, James Bradbury, Gregory
  Chanan, Trevor Killeen, Zeming Lin, Natalia Gimelshein, Luca Antiga, et~al.
\newblock Pytorch: An imperative style, high-performance deep learning library.
\newblock \emph{Advances in neural information processing systems}, 32, 2019.

\bibitem[Pei et~al.(2020)Pei, Wei, Chang, Lei, and Yang]{pei2020geom}
Hongbin Pei, Bingzhe Wei, Kevin Chen-Chuan Chang, Yu~Lei, and Bo~Yang.
\newblock Geom-gcn: Geometric graph convolutional networks.
\newblock \emph{arXiv preprint arXiv:2002.05287}, 2020.

\bibitem[Sun et~al.(2020)Sun, Lin, and Zhu]{sun2020multi}
Ke~Sun, Zhouchen Lin, and Zhanxing Zhu.
\newblock Multi-stage self-supervised learning for graph convolutional networks
  on graphs with few labeled nodes.
\newblock In \emph{Proceedings of the AAAI conference on artificial
  intelligence}, volume~34, pages 5892--5899, 2020.

\bibitem[Sun et~al.(2022)Sun, Li, Peng, Wu, Fu, Ji, and Philip]{sun2022graph}
Qingyun Sun, Jianxin Li, Hao Peng, Jia Wu, Xingcheng Fu, Cheng Ji, and S~Yu
  Philip.
\newblock Graph structure learning with variational information bottleneck.
\newblock In \emph{Proceedings of the AAAI Conference on Artificial
  Intelligence}, volume~36, pages 4165--4174, 2022.

\bibitem[Tang et~al.(2009)Tang, Sun, Wang, and Yang]{tang2009social}
Jie Tang, Jimeng Sun, Chi Wang, and Zi~Yang.
\newblock Social influence analysis in large-scale networks.
\newblock In \emph{Proceedings of the 15th ACM SIGKDD international conference
  on Knowledge discovery and data mining}, pages 807--816, 2009.

\bibitem[Wang(2019)]{wang2019deep}
Minjie~Yu Wang.
\newblock Deep graph library: Towards efficient and scalable deep learning on
  graphs.
\newblock In \emph{ICLR workshop on representation learning on graphs and
  manifolds}, 2019.

\bibitem[Wang et~al.(2021)Wang, Mou, Wang, Xiao, Ju, Shi, and
  Xie]{wang2021graph}
Ruijia Wang, Shuai Mou, Xiao Wang, Wanpeng Xiao, Qi~Ju, Chuan Shi, and Xing
  Xie.
\newblock Graph structure estimation neural networks.
\newblock In \emph{Proceedings of the Web Conference 2021}, pages 342--353,
  2021.

\bibitem[Wieder et~al.(2020)Wieder, Kohlbacher, Kuenemann, Garon, Ducrot,
  Seidel, and Langer]{wieder2020compact}
Oliver Wieder, Stefan Kohlbacher, M{\'e}laine Kuenemann, Arthur Garon, Pierre
  Ducrot, Thomas Seidel, and Thierry Langer.
\newblock A compact review of molecular property prediction with graph neural
  networks.
\newblock \emph{Drug Discovery Today: Technologies}, 37:\penalty0 1--12, 2020.

\bibitem[Wu et~al.(2019)Wu, Wang, Tyshetskiy, Docherty, Lu, and
  Zhu]{wu2019adversarial}
Huijun Wu, Chen Wang, Yuriy Tyshetskiy, Andrew Docherty, Kai Lu, and Liming
  Zhu.
\newblock Adversarial examples on graph data: Deep insights into attack and
  defense.
\newblock \emph{arXiv preprint arXiv:1903.01610}, 2019.

\bibitem[Wu et~al.(2023)Wu, Lin, Liu, Liu, Huang, and Li]{wu2023homophily}
Lirong Wu, Haitao Lin, Zihan Liu, Zicheng Liu, Yufei Huang, and Stan~Z Li.
\newblock Homophily-enhanced self-supervision for graph structure learning:
  Insights and directions.
\newblock \emph{IEEE Transactions on Neural Networks and Learning Systems},
  2023.

\bibitem[Wu et~al.(2022)Wu, Zhao, Li, Wipf, and Yan]{wu2022nodeformer}
Qitian Wu, Wentao Zhao, Zenan Li, David~P Wipf, and Junchi Yan.
\newblock Nodeformer: A scalable graph structure learning transformer for node
  classification.
\newblock \emph{Advances in Neural Information Processing Systems},
  35:\penalty0 27387--27401, 2022.

\bibitem[Xu et~al.(2021)Xu, Xiang, Yu, Cao, and Wang]{xu2021speedup}
Hui Xu, Liyao Xiang, Jiahao Yu, Anqi Cao, and Xinbing Wang.
\newblock Speedup robust graph structure learning with low-rank information.
\newblock In \emph{Proceedings of the 30th ACM International Conference on
  Information \& Knowledge Management}, pages 2241--2250, 2021.

\bibitem[Xu et~al.(2022)Xu, Wu, Liu, Wu, and Wang]{xu2022evidence}
Weizhi Xu, Junfei Wu, Qiang Liu, Shu Wu, and Liang Wang.
\newblock Evidence-aware fake news detection with graph neural networks.
\newblock In \emph{Proceedings of the ACM Web Conference 2022}, pages
  2501--2510, 2022.

\bibitem[Yanardag and Vishwanathan(2015)]{yanardag2015deep}
Pinar Yanardag and SVN Vishwanathan.
\newblock Deep graph kernels.
\newblock In \emph{Proceedings of the 21th ACM SIGKDD international conference
  on knowledge discovery and data mining}, pages 1365--1374, 2015.

\bibitem[Yang et~al.(2016)Yang, Cohen, and Salakhudinov]{yang2016revisiting}
Zhilin Yang, William Cohen, and Ruslan Salakhudinov.
\newblock Revisiting semi-supervised learning with graph embeddings.
\newblock In \emph{International conference on machine learning}, pages 40--48.
  PMLR, 2016.

\bibitem[Yu et~al.(2021)Yu, Zhang, Jiang, Wu, and Yang]{yu2021graph}
Donghan Yu, Ruohong Zhang, Zhengbao Jiang, Yuexin Wu, and Yiming Yang.
\newblock Graph-revised convolutional network.
\newblock In \emph{Machine Learning and Knowledge Discovery in Databases:
  European Conference, ECML PKDD 2020, Ghent, Belgium, September 14--18, 2020,
  Proceedings, Part III}, pages 378--393. Springer, 2021.

\bibitem[Yun et~al.(2019)Yun, Jeong, Kim, Kang, and Kim]{yun2019graph}
Seongjun Yun, Minbyul Jeong, Raehyun Kim, Jaewoo Kang, and Hyunwoo~J Kim.
\newblock Graph transformer networks.
\newblock \emph{Advances in neural information processing systems}, 32, 2019.

\bibitem[Zhang et~al.(2021)Zhang, Zhu, Liu, Wu, Wang, and
  Wang]{zhang2021mining}
Jinghao Zhang, Yanqiao Zhu, Qiang Liu, Shu Wu, Shuhui Wang, and Liang Wang.
\newblock Mining latent structures for multimedia recommendation.
\newblock In \emph{Proceedings of the 29th ACM International Conference on
  Multimedia}, pages 3872--3880, 2021.

\bibitem[Zhang et~al.(2022)Zhang, Gao, Pei, and Huang]{zhang2022robust}
Yanfu Zhang, Hongchang Gao, Jian Pei, and Heng Huang.
\newblock Robust self-supervised structural graph neural network for social
  network prediction.
\newblock In \emph{Proceedings of the ACM Web Conference 2022}, pages
  1352--1361, 2022.

\bibitem[Zhang et~al.(2019)Zhang, Bu, Ester, Zhang, Yao, Yu, and
  Wang]{zhang2019hierarchical}
Zhen Zhang, Jiajun Bu, Martin Ester, Jianfeng Zhang, Chengwei Yao, Zhi Yu, and
  Can Wang.
\newblock Hierarchical graph pooling with structure learning.
\newblock \emph{arXiv preprint arXiv:1911.05954}, 2019.

\bibitem[Zhao et~al.(2021)Zhao, Wang, Shi, Hu, Song, and
  Ye]{zhao2021heterogeneous}
Jianan Zhao, Xiao Wang, Chuan Shi, Binbin Hu, Guojie Song, and Yanfang Ye.
\newblock Heterogeneous graph structure learning for graph neural networks.
\newblock In \emph{Proceedings of the AAAI conference on artificial
  intelligence}, volume~35, pages 4697--4705, 2021.

\bibitem[Zhao et~al.(2023)Zhao, Wen, Ju, Zhang, and Ye]{zhao2023self}
Jianan Zhao, Qianlong Wen, Mingxuan Ju, Chuxu Zhang, and Yanfang Ye.
\newblock Self-supervised graph structure refinement for graph neural networks.
\newblock In \emph{Proceedings of the Sixteenth ACM International Conference on
  Web Search and Data Mining}, pages 159--167, 2023.

\bibitem[Zheng et~al.(2020)Zheng, Zong, Cheng, Song, Ni, Yu, Chen, and
  Wang]{zheng2020robust}
Cheng Zheng, Bo~Zong, Wei Cheng, Dongjin Song, Jingchao Ni, Wenchao Yu, Haifeng
  Chen, and Wei Wang.
\newblock Robust graph representation learning via neural sparsification.
\newblock In \emph{International Conference on Machine Learning}, pages
  11458--11468. PMLR, 2020.

\bibitem[Zhu et~al.(2020)Zhu, Yan, Zhao, Heimann, Akoglu, and
  Koutra]{zhu2020beyond}
Jiong Zhu, Yujun Yan, Lingxiao Zhao, Mark Heimann, Leman Akoglu, and Danai
  Koutra.
\newblock Beyond homophily in graph neural networks: Current limitations and
  effective designs.
\newblock \emph{Advances in Neural Information Processing Systems},
  33:\penalty0 7793--7804, 2020.

\bibitem[Zhu et~al.(2021)Zhu, Xu, Zhang, Du, Zhang, Liu, Yang, and
  Wu]{zhu2021survey}
Yanqiao Zhu, Weizhi Xu, Jinghao Zhang, Yuanqi Du, Jieyu Zhang, Qiang Liu, Carl
  Yang, and Shu Wu.
\newblock A survey on graph structure learning: Progress and opportunities.
\newblock \emph{arXiv e-prints}, pages arXiv--2103, 2021.

\bibitem[Zügner and Günnemann(2019)]{1902.08412}
Daniel Zügner and Stephan Günnemann.
\newblock Adversarial attacks on graph neural networks via meta learning.
\newblock 2019.

\end{thebibliography}

\newpage

\clearpage
\appendix
\begin{appendices}

\clearpage
\section*{Checklist}
\begin{enumerate}

\item For all authors...
\begin{enumerate}
  \item Do the main claims made in the abstract and introduction accurately reflect the paper's contributions and scope?
    \answerYes{}
  \item Did you describe the limitations of your work?
    \answerNA{}.
  \item Did you discuss any potential negative societal impacts of your work?
    \answerNA{}
  \item Have you read the ethics review guidelines and ensured that your paper conforms to them?
    \answerYes{}
\end{enumerate}

\item If you are including theoretical results...
\begin{enumerate}
  \item Did you state the full set of assumptions of all theoretical results?
    \answerNA{}
	\item Did you include complete proofs of all theoretical results?
    \answerNA{}
\end{enumerate}

\item If you ran experiments (e.g. for benchmarks)...
\begin{enumerate}
  \item Did you include the code, data, and instructions needed to reproduce the main experimental results (either in the supplemental material or as a URL)?
    \answerYes{}
  \item Did you specify all the training details (e.g., data splits, hyperparameters, how they were chosen)?
    \answerYes{}
	\item Did you report error bars (e.g., with respect to the random seed after running experiments multiple times)?
    \answerYes{}
	\item Did you include the total amount of compute and the type of resources used (e.g., type of GPUs, internal cluster, or cloud provider)?
    \answerYes{} See \secref{sec:settings}.
\end{enumerate}

\item If you are using existing assets (e.g., code, data, models) or curating/releasing new assets...
\begin{enumerate}
  \item If your work uses existing assets, did you cite the creators?
    \answerYes{}
  \item Did you mention the license of the assets?
    \answerYes{}
  \item Did you include any new assets either in the supplemental material or as a URL?
    \answerNA{}
  \item Did you discuss whether and how consent was obtained from people whose data you're using/curating?
    \answerNA{}
  \item Did you discuss whether the data you are using/curating contains personally identifiable information or offensive content?
    \answerNA{}
\end{enumerate}

\item If you used crowdsourcing or conducted research with human subjects...
\begin{enumerate}
  \item Did you include the full text of instructions given to participants and screenshots, if applicable?
    \answerNA{}
  \item Did you describe any potential participant risks, with links to Institutional Review Board (IRB) approvals, if applicable?
    \answerNA{}
  \item Did you include the estimated hourly wage paid to participants and the total amount spent on participant compensation?
    \answerNA{}
\end{enumerate}

\end{enumerate}

\clearpage
\section{Datasets and Algorithms}

\subsection{Datasets}
\label{appen:dataset}

All of the public datasets used in our benchmark were previously published, either as graph representation learning benchmarks or new datasets for specific graph tasks. The datasets cover various downstream tasks and a multitude of domains: citation network, social network, bioinformatics, website networks, computer vision, and co-occurrence network. \tabref{tab:node-data}, \tabref{tab:hetero-node}, and \tabref{tab:graph-data} provides the detailed  statistics about diverse datasets. We adopt the following benchmark datasets since i) they are widely applied to develop and evaluate GNN models; ii) they contain diverse graph properties from small-scale to large-scale, from homogeneous to heterogeneous, from homophilic to heterophilic, or from node-level to graph-level. The detailed descriptions of these datasets are listed in the following:
\begin{itemize}
    \item \textbf{Cora, Citeseer, Pubmed.} They are the scientific citation network datasets\cite{yang2016revisiting}, where nodes and edges represent the scientific publications and their citation relationships, respectively. Each publication in the dataset is described by a 0/1-valued word vector indicating the absence/presence of the corresponding word from the dictionary. Each node is associated with a one-hot label, where the node classification task is to predict which class the corresponding publication belongs to.

    \item \textbf{Ogbn-arxiv.} The ogbn-arxiv dataset is a benchmark citation network collected in open graph benchmark (OGB) \cite{hu2020open}, which consists of a large number of nodes and edges, and has been widely utilized to evaluate GNN models\footnote{\href{https://ogb.stanford.edu}{https://ogb.stanford.edu}}. Each node represents an arXiv paper from the computer science domain, and each directed edge indicates that one paper cites another one. The node is described by a 128-dimensional word embedding extracted from the title and abstract in the corresponding publication.

    \item \textbf{Cornell, Texas, Wisconsin.} They are the website datasets from WebKB\footnote{\href{http://www.cs.cmu.edu/afs/cs. cmu.edu/project/theo-11/www/wwkb}{http://www.cs.cmu.edu/afs/cs. cmu.edu/project/theo-11/www/wwkb}}, collected from computer science departments of various universities by Carnegie Mellon University\cite{pei2020geom}. While nodes represent webpages in the webpage datasets, edges are hyperlinks between them. The node feature vectors are given by bag-of-word representation of the corresponding webpages. Each node is associated with a one-hot label to indicate one of the following five categories, i.e., student, project, course, staff, and faculty.

    \item \textbf{Actor.} This is an actor co-occurrence network, which is an actor-only induced subgraph of the film-director-actor-writer network\cite{tang2009social}. In the co-occurrence network, nodes correspond to actors and edges denote the co-occurrence relationships on the same Wikipedia pages. Node feature vectors are described by the bag-of-word representation of keywords in the actors' Wikipedia pages.

    \item \textbf{Polblogs.} This is a blog network, which consists of 1,222 vertexes and 16,716 edges. Each node represents a blog page, and each edge denotes a hyperlink between pages. Every blog has a political attribute: conservative or liberal, which is the label of each node.

    \item \textbf{ACM, DBLP, Yelp.} They are real-world heterogeneous graph datasets. DBLP and ACM\cite{yun2019graph} are citation networks, where DBLP contains three types of nodes (papers (P), authors (A), conferences (C)), four types of edges (PA, AP, PC, CP), and research areas of authors as labels; ACM contains three types of nodes (papers (P), authors (A), subjects (S)), four types of edges (PA, AP, PS, SP), and categories of papers as labels. Yelp\cite{lu2019relation} is a review dataset and contains three types of nodes (businesses (B), users (U), services (S)), and 9rating levels (L). The business nodes are labeled by their category.

    \item \textbf{IMDB-B, IMDB-M, REDDIT-B, COLLAB.} IMDB-BINARY and IMDB-MULTI are movie collaboration datasets that consist of the ego-networks of 1,000 actors/actresses who played roles in movies in IMDB. In each graph, nodes represent actors or actresses, and there is an edge between them if they appear in the same movie. REDDIT-BINARY consists of graphs corresponding to online discussions on Reddit. In each graph, nodes represent users, and there is an edge between them if at least one of them responds to the other's comment. COLLAB is a scientific collaboration dataset. Each graph corresponds to a researcher's ego network, i.e., the researcher and its collaborators are nodes and an edge denotes collaboration between two researchers.

    \item \textbf{PROTEINS, MUTAG.} PROTEINS represents macromolecules and it was derived from \citet{dobson2003distinguishing}. The task of it is to predict whether a protein is an enzyme. Nodes represent the amino acids and there is an edge if they are less than 6 Angstroms apart. MUTAG is a collection of nitroaromatic compounds where nodes stand for atoms and edges between nodes represent bonds between the corresponding atoms.

    \item \textbf{Peptides-func, Peptides-struct.} They are proposed by recent long-range graph benchmark\cite{dwivedi2022long} for exploring the ability of GNNs to capture long-range dependencies. Each graph in datasets is a peptide (a short chain of amino acids), while nodes correspond to the heavy (non-hydrogen) atoms and edges represent the bonds between them. Peptides-func is a multi-label graph classification dataset. Graphs are divided into 10 classes based on the peptide functions. Peptides-struct is a multi-label graph regression dataset based on the 3D structure of the peptides.
\end{itemize}

\begin{table}[H]
\caption{Statistics of homogeneous node classification datasets.}
\label{tab:node-data}
\resizebox{1\textwidth}{!}{
\begin{tabular}{lccccccccc}
\toprule
Datasets      & Cora  & Citeseer & Pubmed & ogbn-arxiv & Cornell & Texas & Wisconsin & Actor & Polblogs  \\
\midrule
\#Nodes       & 2,708 & 3,327    & 19,717 & 169,343    & 183     & 183   & 251       & 7,600 & 1,222  \\
\#Edges       & 5,278 & 4,614    & 44,325 & 1,157,799  & 277     & 279   & 450       & 26,659  & 16,716 \\
\#Classes     & 7     & 6        & 3      & 40         & 5       & 5     & 5         & 5  & 2     \\
\#Features    & 1,433 & 3,703    & 500    & 767        & 1,703    & 1,703  & 1,703      & 932  & —   \\
\#Homophily   & 0.81  & 0.74     & 0.80   & 0.65       & 0.12    & 0.06  & 0.18      & 0.22  & 0.91   \\
Avg. \#Degree & 3.90  & 2.77     & 4.50   & 13.67      & 3.03    & 3.05  & 3.59      & 7.02  & 27.36 \\
\bottomrule
\end{tabular}}
\end{table}

\begin{table}[H]
\caption{Statistics of heterogeneous node classification datasets.}
\label{tab:hetero-node}
\begin{tabular}{lccccccccc}
\toprule
Datasets & \#Nodes & \#Edge & \#Node Type & \#Edge Type & \#Features & Avg. \#Degree & \#Training & \#Validation & \#Test \\
\midrule
ACM     & 8,994   & 25,922 & 3           & 4           & 1,902 & 1.51      & 600        & 300          & 2,125  \\
DBLP    & 7,305   & 19,816 & 3           & 4           & 334 & 1.44        & 600        & 300          & 2,057  \\
Yelp    & 3,913   & 77,176 & 3           & 6           & 82 & 3.72         & 300        & 300          & 2,014 \\
\bottomrule
\end{tabular}
\end{table}

\begin{table}[H]
\caption{Statistics of graph-level datasets.}
\label{tab:graph-data}
\resizebox{1\textwidth}{!}{
\begin{tabular}{lcccccccc}
\toprule
Datasets      & IMDB-B & IMDB-M & REDDIT-B & COLLAB & PROTEINS & MUTAG & Peptides-func & Peptides-struct \\ \midrule
\#Graphs     & 1,000   & 1,500   & 2,000     & 5,000   & 1,113     & 188   & 15,535  & 15,535 \\
\#Classes    & 2      & 3      & 2        & 3      & 2        & 2     & 10  & —   \\
Avg. \#Nodes & 19.8   & 13.0   & 429.7    & 74.5   & 39.1     & 17.9  & 150.9  & 150.9 \\ 
Avg. \#Edges & 96.5   & 65.9   & 497.8    & 2,457.5   & 72.8     & 19.8  & 307.3  & 307.3 \\ 
Avg. \#Degree & 4.9   & 5.1   & 1.2    & 33.0   & 1.9     & 1.1  & 2.0  & 2.0 \\ 
Avg. \#Diameter & 1.9   & 1.5    & —    & 1.9    & —     & 8.2   & 27.6  & 27.6 \\ 
Vertex labels & \ding{56}  & \ding{56}  & \ding{56}  & \ding{56}  & \ding{52} & \ding{52} & \ding{56} & \ding{56}
\\
Task  & Class.  & Class.  & Class.  & Class.  & Class.  & Class.  & Class.  & Regre.  \\
Domain & Social & Social & Social & Social & Biochemical & Biochemical & Biochemical & Biochemical \\
\bottomrule
\end{tabular}}
\end{table}

\subsection{Algorithms}
\label{appen:alg}

In our developed GSLB, we integrate 16 state-of-the-art GSL algorithms, including 12 homogeneous node classification models: LDS, GRCN, ProGNN, IDGL, GEN, CoGSL, SLAPS, SUBLIME, STABLE, NodeFormer, GSR, HES-GSL; 2 heterogeneous node classification models: GTN and HGSL; 2 graph-level models: VIB-GSL and HGP-SL. Each is a representative algorithm in its respective task and covers both the early and recent work of GSL. In order to better organize and understand the GSL, we demonstrate a high-level comparison of existing representative algorithms in \tabref{tab:comparison}.

\begin{table}
\renewcommand\arraystretch{1.1}
\caption{Summary of representative Graph Structure Learning (GSL) methods. \emph{Task} refers to the downstream task that the corresponding method is applicable for. For \emph{Oriented}, `Node' means deriving the edge connectivity based on pairwise node embeddings, and `Edge' means directly optimizing the graph adjacency matrix. \emph{Requirement} means the required input data to models for training. For \emph{Graph Regularization}, `SP' means sparsity, `SM' means smoothness, `CON' means connectivity, and `CP' means community preservation.}
\label{tab:comparison}
\begin{tabular}{ccccccccccccc}
\toprule
                               & \multirow{2.5}{*}{Method} & \multirow{2.5}{*}{Task} & \multirow{2.5}{*}{Oriented} & \multicolumn{2}{c}{Requirement}                        &  \multirow{2.5}{*}{Structure Modeling} & \multicolumn{4}{c}{Graph Regularization} & \multirow{2.5}{*}{Complexity} & \multirow{2.5}{*}{Code} \\  \cmidrule(lr){8-11} \cmidrule(lr){5-6}
                               &               &          &                       & Structure & Labels                                                 &                 & SP       & SM       & CON      & CP      &    &                       \\ \midrule
\multirow{9}{*}{\rotatebox{0}{Metric}} 
                               & GRCN \cite{yu2021graph}                   & NC    & Node        & \checkmark       & \checkmark  & Inner product                     & \checkmark         &          &          &         & $O(N^2)$ or $O(NK)$     & \href{https://github.com/PlusRoss/GRCN}{Link}                     \\
                               & IDGL \cite{chen2020iterative}                   & NC  & Node     & \checkmark            & \checkmark    & Cosine similarity                                                     & \checkmark         & \checkmark         & \checkmark         &         & $O(N^2)$ or $O(Nm)$      & \href{https://github.com/hugochan/IDGL}{Link}                    \\
                               & HGSL \cite{zhao2021heterogeneous}                   & HNC   & Node     & \checkmark          & \checkmark    & Cosine similarity                       & \checkmark         &          &          &         & $O(N^2)$    & \href{https://github.com/AndyJZhao/HGSL}{Link}                      \\
                               & HGP-SL \cite{zhang2019hierarchical}                 & GC     & Node         & \checkmark     & \checkmark   & Attention      & \checkmark         &          &          &         & $O(N^2)$      & \href{https://github.com/cszhangzhen/HGP-SL}{Link}                    \\
                               & GEN \cite{wang2021graph}                 & NC      & Node      & \checkmark      & \checkmark       &  Cosine similarity   &                          &          &          &         & $O(N^2)$              & \href{https://github.com/BUPT-GAMMA/Graph-Structure-Estimation-Neural-Networks}{Link}            \\
                               & STABLE \cite{li2022reliable}                 & NC    & Node       & \checkmark        &           & Cosine similarity                   &          &          &          &         & $O(N^2)$              & \href{https://github.com/likuanppd/STABLE}{Link}            \\
                               & NodeFormer \cite{wu2022nodeformer}                 & NC    & Node     &           & \checkmark    & Attention                   & \checkmark         &          &          &         & $O(N)$ or $O(E)$   & \href{https://github.com/qitianwu/NodeFormer}{Link}                       \\ 
                               & GSR \cite{zhao2023self}                 & NC      & Node        & \checkmark     &           & Cosine similarity    &          &          &          &         & $O(NB)$   & \href{https://github.com/andyjzhao/WSDM23-GSR}{Link}                       \\ 
                               & HES-GSL \cite{wu2023homophily}                 & NC      & Node        &     & \checkmark          & Cosine similarity    &          &          &          &         & $O(N^2)$   & \href{https://github.com/LirongWu/Homophil-EnhancedSelf-supervision}{Link}                       \\
                               \midrule
\multirow{6}{*}{\rotatebox{0}{Neural}} & GLCN \cite{jiang2019semi}                   & NC  & Node    &              & \checkmark                                                                            & One-layer neural net                                                   & \checkmark         & \checkmark         &          &         & $O(N^2)$    & \href{https://github.com/jiangboahu/GLCN-tf}{Link}                      \\
                                & PTDNet \cite{luo2021learning}                 & NC    & Node      & \checkmark         & \checkmark                                                                              & Multilayer perceptron               & \checkmark                                            &          &          & \checkmark        & $O(E)$    & \href{https://github.com/flyingdoog/PTDNet}{Link}                      \\
                                & VIB-GSL \cite{sun2022graph}                & GC     & Node        & \checkmark      & \checkmark                                                                                &  Multilayer perceptron                                           & \checkmark         &          &          &         & $O(N^2)$   & \href{https://github.com/RingBDStack/VIB-GSL}{Link}                       \\
                                & NeuralSparse \cite{zheng2020robust}           & NC   & Node       & \checkmark         & \checkmark   & Multilayer perceptron      &          &          &          &         & $O(E)$    & —                      \\
                                & GTN \cite{yun2019graph}                 & HNC     & Edge         & \checkmark     & \checkmark                                                                                & Convolutional layers                                                                       &          &          &          &         & $O(N^2)$      & \href{https://github.com/seongjunyun/Graph_Transformer_Networks}{Link}                    \\
                                & CoGSL \cite{liu2022compact}                  & NC    & Node       & \checkmark        & \checkmark     & Multilayer perceptron                       &          &          &          &         & $O(N^2)$   & \href{https://github.com/liun-online/CoGSL}{Link}                       \\
                                \midrule
\multirow{4}{*}{\rotatebox{0}{Direct}}   & GLNN\cite{gao2020exploring} & NC & Edge & \checkmark & \checkmark & Free variables   & \checkmark & \checkmark & & & $O(N^2)$ & — \\
                                & LDS \cite{franceschi2019learning}                    & NC   & Edge      &           & \checkmark   & Free variables   & \checkmark         &          &          &         & $O(N^2)$   & \href{https://github.com/lucfra/LDS-GNN}{Link}                      \\
                                & LRGNN \cite{xu2021speedup}                  & NC   & Edge        & \checkmark        & \checkmark                                                                                     & Free variables                                                         &          & \checkmark         &          & \checkmark        & $O(N^2)$    & —                      \\
                               & ProGNN \cite{jin2020graph}                 & NC      & Edge      &  \checkmark      & \checkmark                                                                                & Free variables                                                                       & \checkmark         & \checkmark         &          & \checkmark        & $O(N^2)$      & \href{https://github.com/ChandlerBang/Pro-GNN}{Link}                    \\
                               \midrule
\multirow{2}{*}{\rotatebox{0}{Hybird}}  & SLAPS \cite{fatemi2021slaps}                  & NC   & —      &           & \checkmark                                   & Multiple learners                                     &                           &          &          &         & $O(N^2)$    & \href{https://github.com/BorealisAI/SLAPS-GNN}{Link}                      \\
                               & SUBLIME \cite{liu2022towards}                & NC   & —        &         &                                     & Multiple learners  &          &          &          &         & $O(N^2)$   & \href{https://github.com/GRAND-Lab/SUBLIME}{Link}                       \\
                                \bottomrule
\end{tabular}
\end{table}

\begin{itemize}
    \item \textbf{LDS}\cite{franceschi2019learning}. LDS is an early work in GSL. It proposes to approximately solving a bilevel program to jointly learn the graph structure and parameters of GNNs.
    \item \textbf{GRCN}\cite{yu2021graph}. GRCN designs a graph revision module to predict missing edges and revise edge weights. To reduce the complexity of GRCN, Fast-GRCN only calculates the similarity matrix in the first epoch and then computes the values of the kNN-sparse matrix for the remaining epochs. Therefore, the complexity of GRCN can be reduced to $O(NK)$, where $K$ is the number of top-$K$ important neighbors of each node.
    \item \textbf{ProGNN}\cite{jin2020graph}. ProGNN treats the adjacency matrix as a learnable variable and directly optimizes it with GNNs to learn a robust structure.
    \item \textbf{IDGL}\cite{chen2020iterative}. IDGL jointly and iteratively learns graph structures and graph embeddings. In addition, it also provides a scalable version, namely IDGL-ANCH, which randomly samples $m$ anchors from the node set for each node to calculate affinity scores.
    \item \textbf{GEN}\cite{wang2021graph}. GEN designs a structure model characterizing the underlying graph generation and an observation model injecting multi-order neighborhood information to accurately infer the graph structure based on Bayesian inference.
    \item \textbf{CoGSL}\cite{liu2022compact}. CoGSL utilizes mutual information compression to extract compact and robust graph structure, namely "minimal sufficient structure", which maximizes the information about downstream tasks.
    \item \textbf{SLAPS}\cite{fatemi2021slaps}. SLAPS focuses on topology inference task and provides more supervision signals for inferring a graph structure through self-supervision learning.
    \item \textbf{SUBLIME}\cite{liu2022towards}. To prevent the reliance on labels, bias of edge distribution, and the limitation on application tasks, SUBLIME learns graph structure in an unsupervised manner. It first generates a learning target graph, anchor graph, and maximizes the agreement between the anchor graph and the learned graph by contrastive learning. It also designs a variety of graph learners and post-processors.
    \item \textbf{STABLE}\cite{li2022reliable}. STABLE optimizes the learned graph structure through an unsupervised pipeline to avoid using unreliable supervision signals.
    \item \textbf{NodeFormer}\cite{wu2022nodeformer}. NodeFormer introduces a novel all-pair message-passing scheme for efficiently propagating node signals between arbitrary nodes. Because of the high complexity of Transformer-style architecture, it uses a kernelized Gumbel-Softmax operator to reduce the complexity to linearity.
    \item \textbf{GSR}\cite{zhao2023self}. GSR first estimates the underlying graph structure by a multi-view contrastive learning framework and then fine-tunes a GNN on the learned structure.
    \item \textbf{HES-GSL}\cite{wu2023homophily}. HES-GSL proves that the task-specific supervision signals may be insufficient to support the learning of both graph structure and parameters of GNNs. Therefore, it proposes homophily-enhanced self-supervision for GSL to provide more supervision information for topology inference.
    \item \textbf{GTN}\cite{yun2019graph}. GTN learns a soft selection of edge types and composite relations for generating useful multi-hop connections for heterogeneous graphs.
    \item \textbf{HGSL}\cite{zhao2021heterogeneous}. HGSL firstly attempts to learn heterogeneous graph structure and GNNs jointly. It considers the feature similarity by generating a feature similarity graph and optimizes the complex heterogeneous interactions by generating a feature propagation graph and semantic graph.
    \item \textbf{HGP-SL}\cite{zhang2019hierarchical}. HGP-SL is a graph-level model. It adaptively selects a subset of nodes to form an induced subgraph and utilizes structure learning to refine subgraphs at each layer.
    \item \textbf{VIB-GSL}\cite{sun2022graph}. VIB-GSL firstly attempts to advance the Information Bottleneck (IB) principle for graph structure learning and proposes to use dot-product self-attention to refine dynamic connections.
\end{itemize}

\section{Additional Experimental Results}
\label{appen:exp}

\subsection{Performance on long-range datasets}
Whether GSL can capture long-range dependencies is an interesting topic. Traditional message-passing GNNs simply rely on local neighbors to produce node representations and are hard to learn higher-order information. Recently, LRGB\cite{dwivedi2022long} presents a series of graph learning datasets, which arguably require the ability of long-range interactions to achieve strong performance. \tabref{tab:long} shows the experimental results of graph-level GSL algorithms on two long-range datasets. We can observe that VIB-GSL and HGP-SL do not exhibit promising results on \texttt{Peptides-Func} dataset, but show evidently better results on \texttt{Peptides-Struct} dataset than the baseline model. We suspect that this may be related to whether the GSL algorithm is suitable for specific downstream tasks. We only investigate the long-range capability of graph-level GSL models. The investigation of long-range capability at the node-level is our future work.

\begin{table}[t]
\caption{Average Precision (AP) $\pm$ STD comparison (\%) for \texttt{Peptides-func} and Mean Absolute Error (MAE) $\pm$ STD comparison for \texttt{Peptides-struct}. Each result was obtained from 3 repeated experiments with different random seeds. $\uparrow$ represents the larger, the better while $\downarrow$ represents the smaller, the better.}
\label{tab:long}
\begin{center}
\begin{tabular}{lcc}
\toprule
Method         & \texttt{Peptides-Func} (AP $\uparrow$) & \texttt{Peptides-Struct} (MAE $\downarrow$) \\
\midrule
GCN            & 58.31{\tiny$\pm$0.28}         & 0.3566{\tiny$\pm$0.0014}          \\
GAT            & 45.59{\tiny$\pm$0.35}         & 0.4375{\tiny$\pm$0.0036}          \\
GIN            & \best\textbf{60.92{\tiny$\pm$0.88}}         & 0.3647{\tiny$\pm$0.0028}          \\
SAGE           & \second\underline{59.41{\tiny$\pm$0.57}}         & 0.3716{\tiny$\pm$0.0011}          \\
VIB-GSL (GCN)  & 52.05{\tiny$\pm$2.40}         & 0.3089{\tiny$\pm$0.0036}          \\
VIB-GSL (GAT)  & 42.64{\tiny$\pm$0.33}         & 0.3992{\tiny$\pm$0.0129}             \\
VIB-GSL (GIN)  & 47.61{\tiny$\pm$0.93}         & 0.3032{\tiny$\pm$0.0011}          \\
VIB-GSL (SAGE) & 54.11{\tiny$\pm$0.74}         & 0.3063{\tiny$\pm$0.0014}          \\
HGP-SL (GCN)         & 52.16{\tiny$\pm$0.94}         & 0.2935{\tiny$\pm$0.0104}          \\
HGP-SL (GIN)         & 55.30{\tiny$\pm$0.88}         & \best\textbf{0.2786{\tiny$\pm$0.0018}}          \\
HGP-SL (SAGE)         & 53.48{\tiny$\pm$0.71}         & \second\underline{0.2794{\tiny$\pm$0.0058}}          \\
\bottomrule
\end{tabular}
\end{center}
\end{table}

\subsection{Visualiuzation}
\label{appen:visual}

\begin{figure}[t]
\centering
\includegraphics[width=\textwidth]{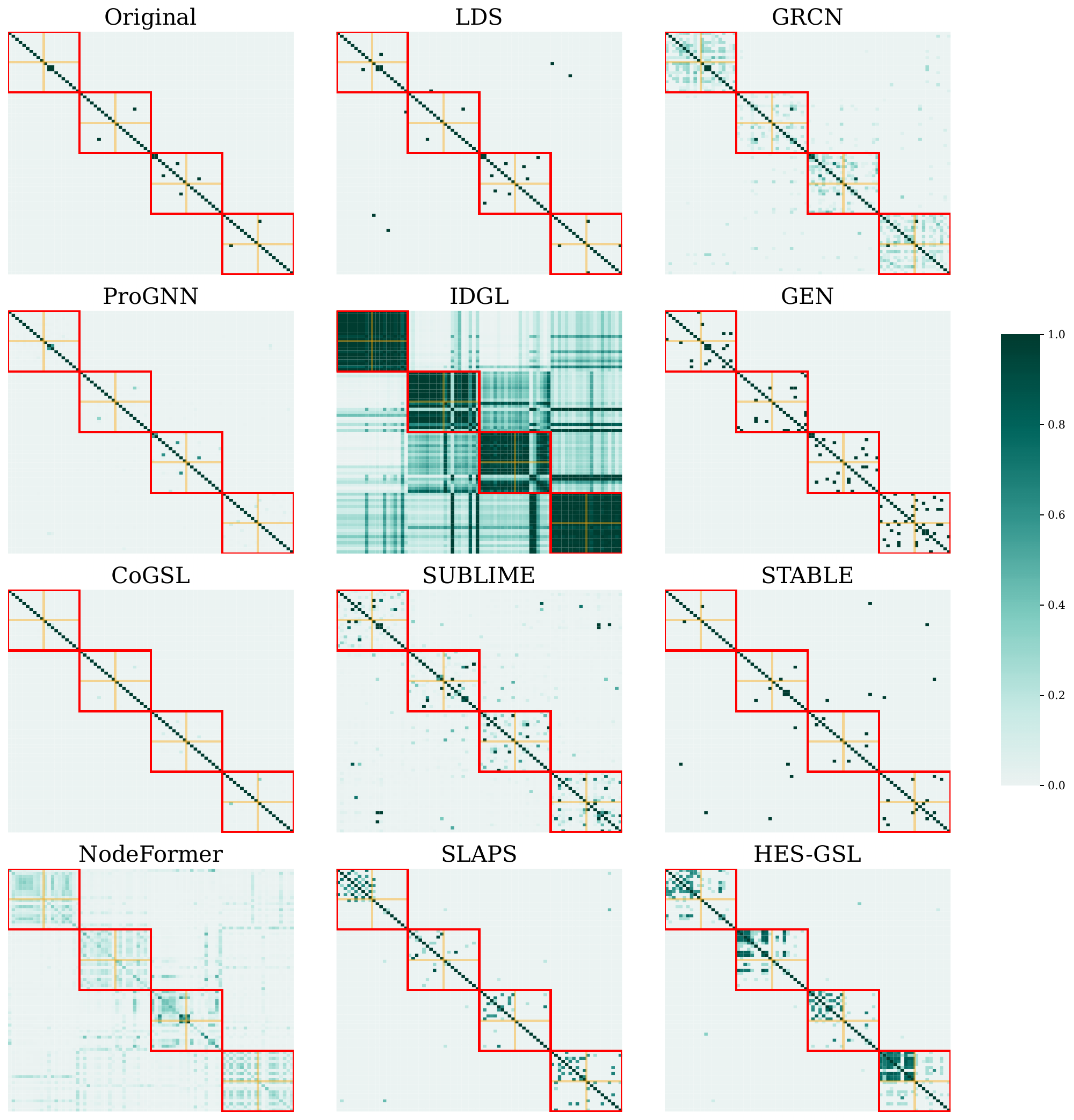}
\caption{Visualization of the original graph of Cora and the learned graphs by various GSL algorithms.}
\label{fig:visualization}
\end{figure}

\begin{figure}[t]
\centering
\includegraphics[width=\textwidth]{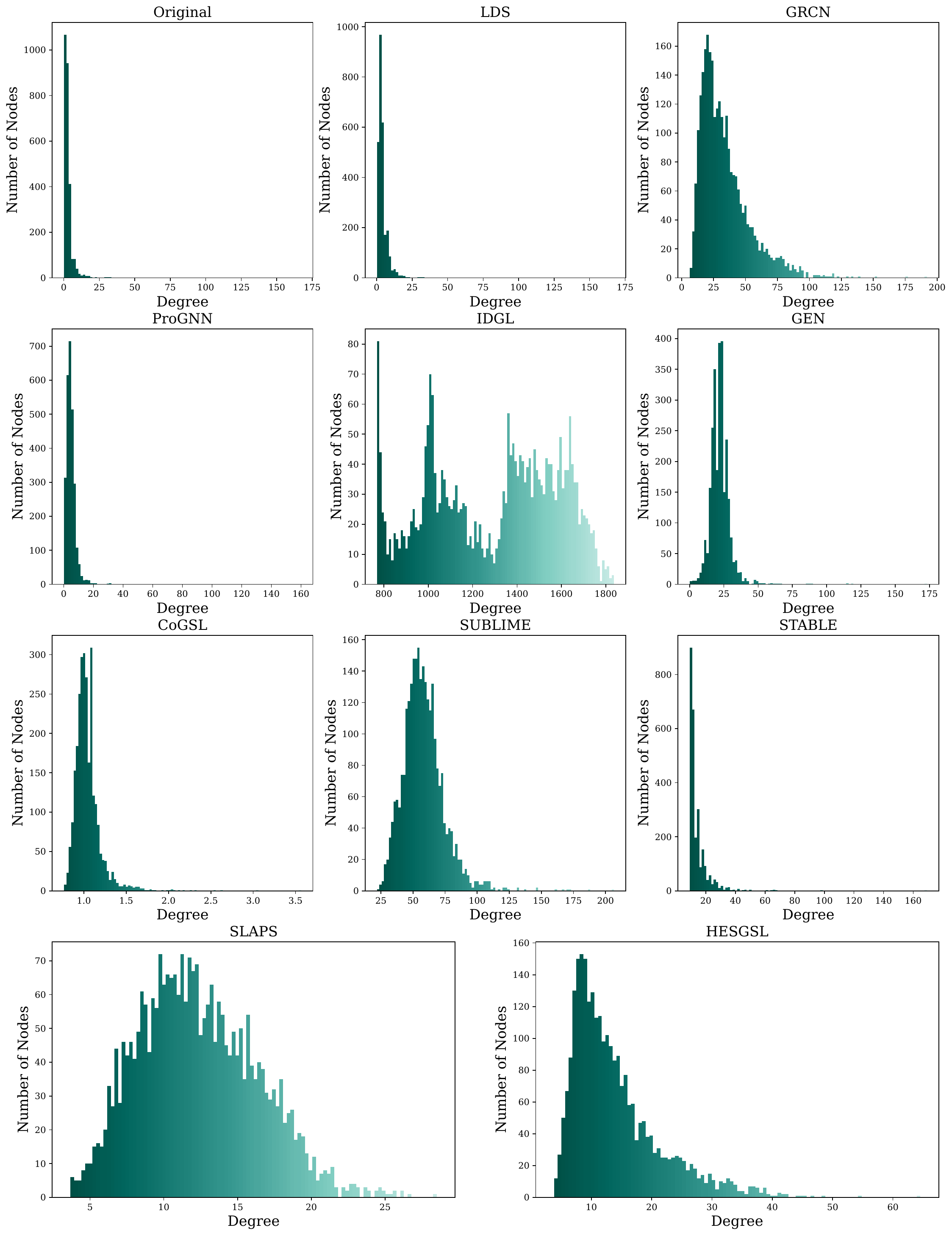}
\caption{Degree distribution of the original graph of Cora and the learned graphs by various GSL algorithms.}
\label{fig:degree}
\end{figure}

In order to more intuitively understand the ability and characteristics of GSL, we visualize the original graph structure of Cora and the learned graphs of various GSL algorithms in \figref{fig:visualization}. We select four categories of Cora and randomly sample 10 labeled nodes and 10 unlabeled nodes to extract a subgraph. The elements inside the red rectangle are the intra-class connections, and the elements on the diagonal are self-loops. We can observe that i) most of the learned structures are much denser than the original structure (especially the learned structure of IDGL); ii) in the TI scenarios, SLAPS and HES-GSL will prefer to connect labeled nodes. Moreover, we also make statistics on the properties of the original structure and the learned structures. As shown in \tabref{tab:property}, we list node homophily ratio\cite{pei2020geom}, edge homophily ratio\cite{zhu2020beyond}, class insensitive edge homophily ratio\cite{lim2021large} of the structures. They can be calculated as follows:

\begin{equation}
    \text{Node: }\ \ \ \ \ h_{node}=\frac{1}{N}\sum_{v\in\mathcal{V}}\frac{\{(u,v):u\in\mathcal{N}(v)\land y_v=y_u\}}{|\mathcal{N}(v)|}
    \nonumber
\end{equation}
\begin{equation}
    \text{Edge: }\ \ \ \ \ h_{edge}=\frac{|\{(v,u):(v,u)\in\mathcal{E}\land y_v=y_u\}|}{|\mathcal{E}|}
    \nonumber
\end{equation}
\begin{equation}
    \text{EI: }\ \ \ \ \ h_{ei}=\frac{1}{C-1}\sum_{k=1}^C\max\bigg(0, h_{edge}^k-\frac{|\mathcal{C}_k|}{|\mathcal{V}|}\bigg)
    \nonumber
\end{equation}

where $\mathcal{V}$ and $\mathcal{E}$ denotes the set of nodes and edges, $C$ is the number of classes, $|\mathcal{C}_k|$ denotes the number of nodes of class $k$, and $h_{edge}^k$ denotes the edge homophily ratio of class $k$. And we also present the degree assortativity coefficient\cite{newman2003mixing} of structures, which refers to the tendency of nodes to connect with other similar nodes over dissimilar nodes. According to \tabref{tab:property}, we can observe that the learned graph structures do not increase the homophily ratio, thus the homophily ratio may not be the main reason for the improved performance of GSL. We also find that all GSL algorithms make the graph structure mode dense, which indicates that real-world datasets may be too sparse. Finally, we plot the degree distribution of the original graph of Cora and the learned graphs by various GSL algorithms in \figref{fig:degree}. We can observe that the learned graph structure still follows the long-tail distribution. How to use the GSL to alleviate the unfairness of the classification performance of nodes with different degrees is also a problem worth exploring.

\begin{table}[t]
\caption{Graph property statistics of the original and the learned graphs on Cora.\emph{Node Homo.} represents node homophily ratio; \emph{Edge Homo.} represents edge homophily ratio; \emph{EI Homo.} represents class insensitive edge homophily ratio; \emph{Assor.} represents the degree assortativity coefficient.}
\label{tab:property}
\begin{center}
\begin{tabular}{lccccc}
\toprule
Method     & Node Homo. & Edge Homo. & EI Homo. & Assor.  & Density \\
\midrule
Original   & 0.8252     & 0.8100     & 0.7657   & -0.0659 & 0.0014  \\
LDS        & 0.7636     & 0.7500     & 0.6842   & -0.0571 & 0.0019  \\
GRCN       & 0.5948     & 0.5941     & 0.4857   & -0.0720 & 0.0816  \\
ProGNN     & 0.2208     & 0.2041     & 0.0309   & -0.0246 & 0.5666  \\
IDGL       & 0.1796     & 0.1796     & 0.0000   & /       & 1.0000  \\
GEN        & 0.7425     & 0.7612     & 0.7055   & 0.0560  & 0.0085  \\
CoGSL      & 0.7300     & 0.7081     & 0.6343   & 0.0469  & 0.0180  \\
SUBLIME    & 0.1796     & 0.1796     & 0.0000   & /       & 1.0000  \\
STABLE     & 0.5259     & 0.5228     & 0.4049   & -0.0153 & 0.0057  \\
NodeFormer & 0.1796     & 0.1796     & 0.0000   & /       & 1.0000  \\
SLAPS      & 0.5683     & 0.5755     & 0.4820   & 0.0709  & 0.0107  \\
HESGSL     & 0.6353     & 0.6635     & 0.5835   & 0.0989  & 0.0171  \\
\bottomrule
\end{tabular}
\end{center}
\end{table}

\end{appendices}

\end{document}